\documentclass{article}

\PassOptionsToPackage{numbers, compress}{natbib}

\usepackage[final]{neurips_2025}

\usepackage[utf8]{inputenc}
\usepackage[T1]{fontenc}
\usepackage{hyperref}
\usepackage{url}
\usepackage{booktabs}
\usepackage{amsfonts}
\usepackage{nicefrac}
\usepackage{microtype}
\usepackage[table]{xcolor}
\usepackage[T1]{fontenc} 
\usepackage{acronym}
\usepackage{amsmath}
\usepackage{multirow}
\usepackage{siunitx}
\usepackage{threeparttable}
\usepackage{graphicx}
\usepackage{subcaption}
\usepackage{wrapfig}
\usepackage{threeparttable}

\newacro{SOTA}[SOTA]{state-of-the-art}
\newacro{RF}[RF]{radio frequency}
\newacro{FMCW}[FMCW]{frequency modulated continuous wave}
\newacro{GCN}[GCN]{graph convolutional network}
\newacro{PC}[PC]{point cloud}
\newacro{MLP}[MLP]{multiple layer perceptron}
\newacro{GT}[GT]{ground truth}
\newacro{BBox}[BBox]{bounding box}

\newcommand{\ba}{\begin{array}}
\newcommand{\ea}{\end{array}}
\newcommand{\be}{\begin{displaymath}}
\newcommand{\ee}{\end{displaymath}}
\newcommand{\ben}{\begin{equation}}
\newcommand{\een}{\end{equation}}
\newcommand{\bena}{\begin{eqnarray}}
\newcommand{\eena}{\end{eqnarray}}
\newcommand{\beqa}{\begin{eqnarray*}}
\newcommand{\enqa}{\end{eqnarray*}}

\newcommand{\bc}{\begin{center}}
\newcommand{\ec}{\end{center}}
\newcommand{\bi}{\begin{itemize}}
\newcommand{\ei}{\end{itemize}}
\newcommand{\benu}{\begin{enumerate}}
\newcommand{\eenu}{\end{enumerate}}
\newcommand{\bdes}{\begin{description}}
\newcommand{\edes}{\end{description}}
\newcommand{\bt}{\begin{tabular}}
\newcommand{\et}{\end{tabular}}

\newcommand \bbf{{\bf b}}
\newcommand \cbf{{\bf c}}

\newcommand \fbf{{\bf f}}
\newcommand \gbf{{\bf g}}

\newcommand \lbf{{\bf l}}

\newcommand \pbf{{\bf p}}
\newcommand \qbf{{\bf q}}

\newcommand \tbf{{\bf t}}

\newcommand \Bbf{{\bf B}}

\newcommand \Fbf{{\bf F}}

\newcommand \Kbf{{\bf K}}

\newcommand \Mbf{{\bf M}}

\newcommand \Pbf{{\bf P}}
\newcommand \Qbf{{\bf Q}}
\newcommand \Rbf{{\bf R}}

\newcommand \Tbf{{\bf T}}

\newcommand \Wbf{{\bf W}}

\newcommand \Ybf{{\bf Y}}
\newcommand \Zbf{{\bf Z}}

\newcommand{\Rset}{{\mathbb R}}

\newcommand{\circlambda}{\mbox{$\Lambda$
             \kern-.85em\raise1.5ex
             \hbox{$\scriptstyle{\circ}$}}\,}

\definecolor{deepskyblue}{RGB}{54, 125, 189}
\definecolor{lightskyblue}{RGB}{58, 178, 198}
\hypersetup{colorlinks = true, linkcolor = deepskyblue,
            urlcolor  = lightskyblue,
            citecolor = deepskyblue,
            anchorcolor = deepskyblue}

\title{RAPTR: Radar-based 3D Pose Estimation \\using Transformer}

\author{%
    Sorachi Kato$^{1,2}$\thanks{The work was initiated during his internship at MERL.}, \, Ryoma Yataka$^{1, 3}$, \, Pu (Perry) Wang$^{1}$\thanks{Project Lead.}, \
    Pedro Miraldo$^{1}$, \\ \textbf{Takuya Fujihashi$^{2}$,} \, \textbf{Petros Boufounos$^{1}$} \\
    $^{1}$Mitsubishi Electric Research Laboratories~(MERL), USA \\
    $^{2}$The University of Osaka, Japan \\
    $^{3}$Information Technology R\&D Center (ITC), Mitsubishi Electric Corporation, Japan
}

\begin{document}
\maketitle

\begin{abstract}
Radar-based indoor 3D human pose estimation typically relied on fine-grained 3D keypoint labels, which are costly to obtain especially in complex indoor settings involving clutter, occlusions, or multiple people. In this paper, we propose \textbf{RAPTR} (RAdar Pose esTimation using tRansformer) under weak supervision, using only 3D BBox and 2D keypoint labels which are considerably easier and more scalable to collect. 
Our RAPTR is characterized by a two-stage pose decoder architecture with a pseudo-3D deformable attention to enhance (pose/joint) queries with multi-view radar features: a pose decoder estimates initial 3D poses with a 3D template loss designed to utilize the 3D BBox labels and mitigate depth ambiguities; and a joint decoder refines the initial poses with 2D keypoint labels and a 3D gravity loss.
Evaluated on two indoor radar datasets, RAPTR outperforms existing methods, reducing joint position error by $34.3\%$ on HIBER and $76.9\%$ on MMVR. Our implementation is available at \url{https://github.com/merlresearch/radar-pose-transformer}.
\end{abstract}

\section{Introduction}
\label{sec:intro}

Accurate human perception is essential for indoor applications, including elderly monitoring, smart building management, and robotic navigation.
Although vision sensors offer high spatial resolution, they raise privacy concerns and perform poorly under low light, occlusions, and hazardous conditions (fire or smoke). 
In contrast, radar provides penetration capability, robustness to adverse conditions, and low deployment cost, ideal for privacy-preserving indoor sensing~\cite{Yao2024, Lu20a, Pandharipande2023_SensingMachineLearning, Skog2024_humandetection4dradar, MLRF25}.

 By processing 4D radar tensors, RF-Pose 3D~\cite{zhao2018rfpose3d} demonstrated through-the-wall 3D pose estimation with a convolutional neural network~(CNN), while HRRadarPose~\cite{yuan2024rtpose} employed an hourglass neural network HRNet~\cite{Wang2021_HRNet}. 
 mRI~\cite{Sizhe2022_mRI} is a multi-modal 3D human pose estimation dataset that integrates mmWave radar, RGB-D cameras, and inertial sensors to facilitate research in human pose estimation and action detection.
 QRFPose~\cite{wan2024qrfpose} is a novel approach that adopts a DETR~\cite{carion2020detr}-style query mechanism for end-to-end 3D regression using multi-view radar perceptions.
Existing pipelines often rely on expensive fine-grained 3D keypoint labels \cite{RFMask23}, typically collected using non-portable 3D motion capture systems such as VICON, or using LiDAR, which can still suffer from occlusions and incomplete observations.

Collecting cheaper, lower-cost labels, such as fine-grained 2D keypoints in the image plane and/or coarse-grained 3D bounding boxes (BBoxes), is considerably easier and more scalable particularly in complex indoor settings (e.g., cluttered, occlusion, multi-person), compared with acquiring dense 3D keypoints labels. Examples include RF-Pose \cite{zhao2018through}, HuPR \cite{lee2023hupr}, and, more recently, MMVR \cite{Rahman2024_mmvr} datasets. To the best of our knowledge, the use of 2D keypoints and 3D BBoxes, as a substitute for costly 3D keypoints, for radar-based 3D human pose estimation has not been systematically investigated in the literature before. 

\begin{figure*}[t]
    \centering
    \includegraphics[width=\linewidth]{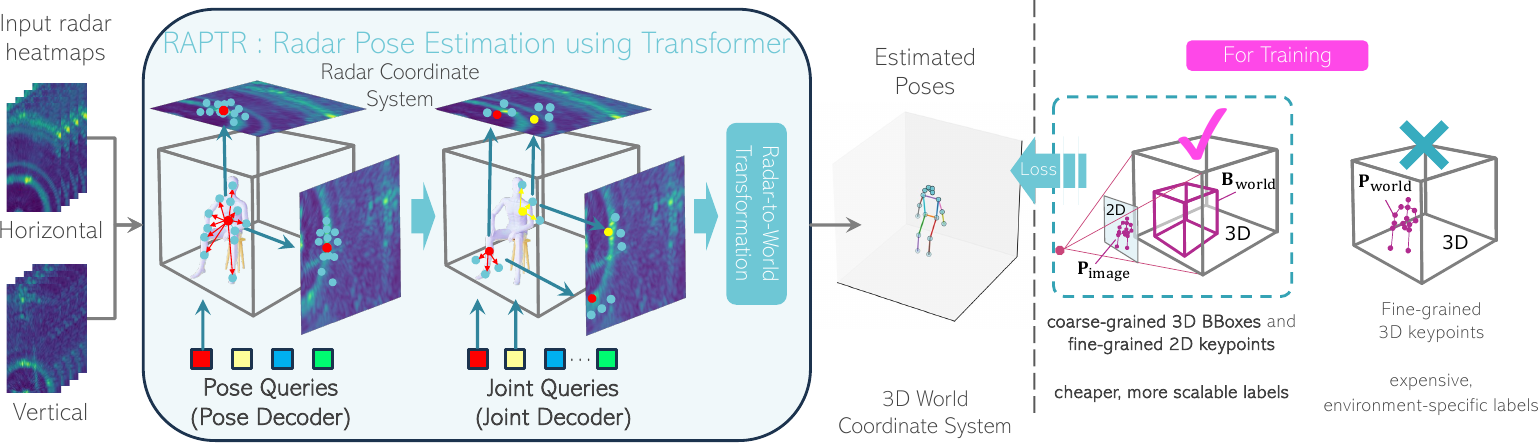}
    \vspace{-3mm}
    \caption{
    {RAPTR}  takes multi-view radar heatmaps as inputs and performs a novel Pseudo-3D deformable attention between (pose and joint) queries and multi-view radar features in a two-stage decoder to estimate 3D human poses in a 3D coordinate system. Rather than relying on expensive, environment‐specific fine‐grained 3D keypoint labels, RAPTR makes use of cheaper, more scalable labels such as coarse-grained 3D BBoxes and fine-grained 2D keypoints to train the model. 
    }
    \label{fig:concept}
    \vspace{-2mm}
\end{figure*}

To address this gap, we propose \textbf{RAPTR} (RAdar Pose esTimation using tRansformer) in Fig.~\ref{fig:concept}, a radar-based pipeline designed to take multi-view radar heatmaps as inputs and estimate 3D human poses under weak supervision using only 3D BBox and 2D keypoint labels.
RAPTR builds on the two-stage (pose and joint) decoder architecture of the state-of-the-art RGB-based 2D pose estimation PETR framework \cite{shi2022_petr} and introduces a structural loss function that is designed to utilize weak supervision labels to mitigate the depth ambiguity. 
RAPTR also lifts the 2D deformable attention in PETR to a pseudo-3D deformable attention, wherein reference points (dots in Fig.~\ref{fig:concept}) and offsets (arrows in Fig.~\ref{fig:concept}) are proposed in the 3D radar coordinate system and projected onto multiple radar views (dots on the radar heatmaps in Fig.~\ref{fig:concept}) to eliminate redundant per-view offset estimation and offer better scalability as the number of radar views increases.
Our model outperforms a list of radar-based 3D pose baselines over two indoor radar datasets: HIBER~\cite{RFMask23} and MMVR~\cite{Rahman2024_mmvr}.
The main contributions of this work are: 
\begin{itemize}
    \item To the best of our knowledge, RAPTR is the first radar-based 3D human pose estimation framework to explicitly utilize low-cost weak supervision in the form of 3D BBoxes and 2D keypoints, rather than relying on fine-grained 3D keypoint labels. 
    \item We introduce a structured loss function that tightly couples the two-stage decoder architecture to enable 3D pose estimation under weak supervision. Specifically, we design a 3D Template Loss, which utilizes the 3D BBox labels at the pose decoder, and a combined 3D Gravity and 2D Keypoint Loss at the pose decoder, allowing RAPTR to effectively learn geometrically consistent 3D poses from weak supervision.
    \item We further introduce a pseudo-3D deformable attention mechanism to bridge the 3D spatial domain and 2D radar views, enabling scalable view association while preserving pose estimation performance. 
\end{itemize}

\section{Related Work}
\label{sec:related_works}

\paragraph{Human Pose Estimation with RGB Image:}
Human pose estimation from images involves localizing body joints for multiple subjects and associating them for each subject. Existing architectures fall into two main paradigms: top-down and bottom-up.
The top-down methods first detect each person using detectors such as Faster R-CNN~\cite{ren2017faster} or Mask R-CNN~\cite{he2020mask}, then applying a single-person pose estimator to each cropped region. These approaches achieve state-of-the-art accuracy with models like Stacked-Hourglass~\cite{newell2016stacked}, HRNet~\cite{sun2019deep}, and DarkPose~\cite{zhang2020distribution-aware}.
In contrast, bottom-up methods such as OpenPose~\cite{cao2017realtime}, HigherHRNet~\cite{cheng2020higherhrnet}, and SAHR~\cite{luo2021rethinking} bypass the detection step by predicting all joint candidates across the entire image and grouping them into individuals. PETR~\cite{shi2022_petr} introduces an end-to-end pose estimation framework using a query-based, two-stage transformer decoder architecture.
Beyond 2D, recent methods addresses 3D pose from RGB or RGB-D inputs, either by directly regressing 3D joints~\cite{lutz2022jointformer} or by lifting 2D predictions into the 3D space  through geometric reasoning or weak supervision~\cite{chen20173d, chen2019unsupervised, martinez2017simple, tome2017lifting}.

\paragraph{Human Pose Estimation with radar or radio frequency signals:}
Recent studies have shown that information extracted from commercial radars is sufficiently informative to perform fine-grained human pose estimation, both for 2D and 3D.
Despite the coarse-grained nature of the radar point clouds (PCs), deep neural pipelines have achieved a multitude of performance gains~\cite{RadHAR19,mmPose20,Xue2021_mmMesh,Sizhe2022_mRI,mmFi23, mmPoint23, Hu2024_mmPoseFK, Kini2024_TransHuPR, xue2022m4esh, yang2024mmbat, fan2024diffusion,milliFlow, yang2025mmdear}.
On the other hand, methods using raw radar measurements and radar heatmaps have been widely explored~\cite{zhao2018through,CubeLearn23, lee2023hupr,RFMask23, yuan2024rtpose,Rahman2024_mmvr, wan2024qrfpose, yataka2024retr, mmDiffusion25}.
RF-Pose~\cite{zhao2018through} pioneered multi-view 3D CNNs for through-wall 2D estimation. HuPR~\cite{lee2023hupr} refines such heatmaps via a \ac{GCN}. HRRadarPose~\cite{yuan2024rtpose} adopts an HRNet-style~\cite{Wang2021_HRNet} single-stage head for 3D output.
QRFPose~\cite{wan2024qrfpose}, based on a DETR-style Transformer~\cite{carion2020detr} for end-to-end query-based 3D pose estimation, is the closest baseline to ours. It differs by applying per-view 2D deformable attention and using a single decoder for all keypoints, followed by grouping. In contrast, our method employs pseudo-3D deformable attention and a two-stage decoder.

\section{Preliminary}
\label{sec:Preliminary}
\begin{wrapfigure}[13]{r}{2.1in}
    \vspace{-0.2in}
    \centering
    \includegraphics[width=0.8\linewidth]{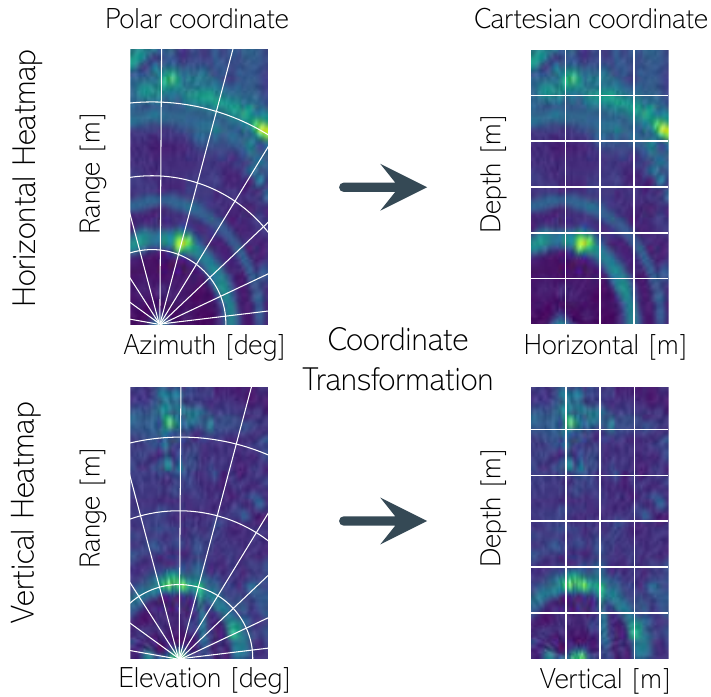}
    \vspace{-2mm}
    \caption{Multi-view radar heatmaps.}
    \label{fig:radar}
\end{wrapfigure}

\paragraph{Multi-View Radar Heatmaps:}
As shown in Fig.~\ref{fig:radar}, two synchronized radar arrays (horizontal and vertical) collect reflected pulses that form a 3D data cube per array (ADC samples $\times$ pulses $\times$ elements). A 3D FFT converts each cube into a range–Doppler–angle spectrum, whose angle dimension is azimuth for the horizontal array and elevation for the vertical array. After Doppler-axis integration to boost SNR (signal-to-noise ratio), we obtain two polar 2D heatmaps (range-azimuth and range-elevation).
These are mapped to the Cartesian space: $\Ybf_{\mathtt{hor}}(t)\in\mathbb{R}^{W\times D}$ for horizontal-depth and $\Ybf_{\mathtt{ver}}(t)\in\mathbb{R}^{H\times D}$ for vertical-depth at frame $t$.
The temporal context is captured by stacking $T$ consecutive frames, giving $\Ybf_{\mathtt{hor}}\in\mathbb{R}^{T\times W\times D}$ and $\Ybf_{\mathtt{ver}}\in\mathbb{R}^{T\times H\times D}$.

\paragraph{Problem Formulation:}
The 3D pose estimation task takes $T$ consecutive radar frames, $\Ybf_{\mathtt{hor}}$ and $\Ybf_{\mathtt{ver}}$, as input and estimates poses $\hat{\Pbf}_\mathtt{world}$ in the 3D world coordinate system, 
\begin{align}
    \hat{\Pbf}_\mathtt{world} &= \mathcal{T}_\mathtt{r2w}(\hat{\Pbf}_\mathtt{radar}) = \mathcal{T}_\mathtt{r2w}(f(\Ybf_{\mathtt{hor}}, \Ybf_{\mathtt{ver}})),
\end{align}
where $f$ represents the 3D pose estimation pipeline in the 3D radar coordinate system, and $\mathcal{T}_\mathtt{r2w}$ is a known radar-to-world coordinate transformation that converts the estimated 3D poses into the 3D world coordinate system.
Rather than relying on costly, non-scalable fine‐grained 3D keypoint labels $\Pbf_{\mathtt{world}}$, we consider cheaper, more scalable labels such as coarse-grained 3D BBoxes $\Bbf_{\mathtt{world}}$ and fine-grained 2D keypoints $\Pbf_{\mathtt{image}}$ for supervision, as shown in Fig.~\ref{fig:concept}. 

\section{RAPTR: Radar-based 3D Pose Estimation using Transformer}
\label{sec:methodology}
\begin{figure*}
    \centering
    \includegraphics[width=\linewidth]{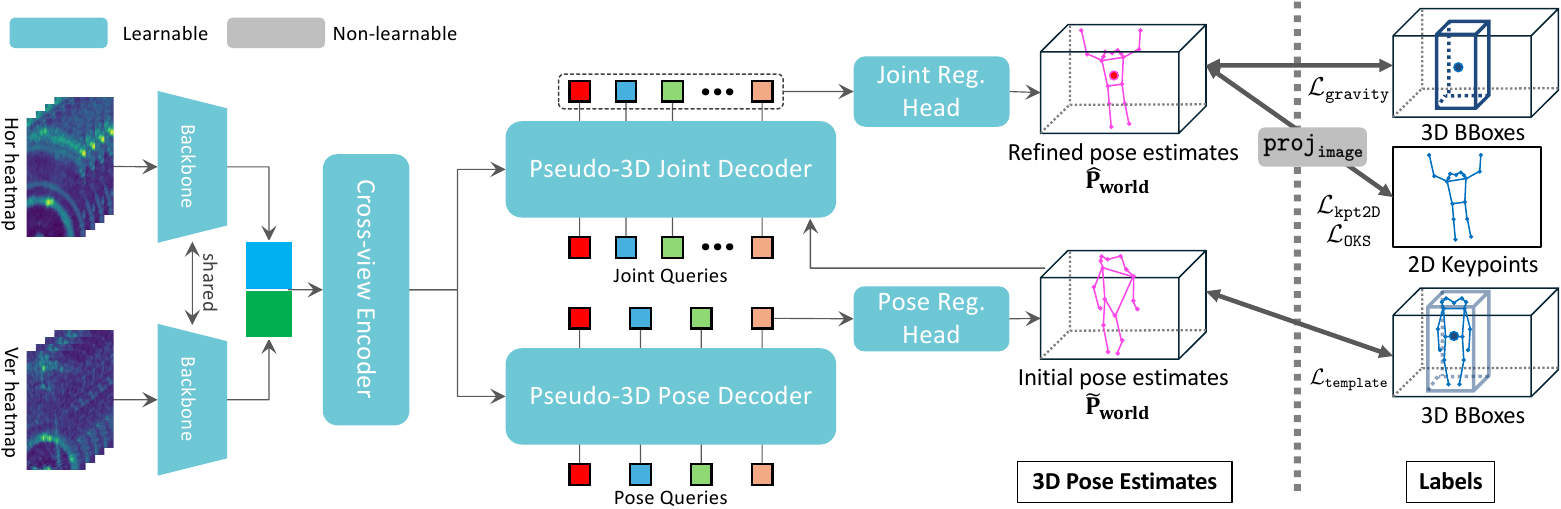}
    \caption{
    The  RAPTR architecture consists of: 1) \textbf{Cross-view Encoder} that extracts multi-scale radar features; 2) \textbf{Pseudo-3D Pose Decoder} that enhances pose queries via a pseudo-3D deformable attention and predicts initial 3D poses; and 3) \textbf{Pseudo-3D Joint Decoder} that further refines joint queries and outputs final 3D poses. In terms of \textbf{loss function}, RAPTR leverages 3D BBox and 2D keypoint labels through coarse-grained 3D loss (gravity and template) and 2D keypoint loss. 
    }
    \label{fig:architecture}
    \vspace{-2mm}
\end{figure*}

We present the RAPTR architecture in Fig.~\ref{fig:architecture}, following a left-to-right order, and highlight radar-specific modifications. Refer to Appendix~\ref{app:network_architecture} for detailed architecture and computational complexity.

\subsection{Architecture}

\textbf{Backbone}: Given $\Ybf_{\mathtt{hor}}\in {\mathbb{R}}^{T\times W\times D}$ and $\Ybf_{\mathtt{ver}}\in {\mathbb{R}}^{T\times H\times D}$, a shared backbone network (e.g., ResNet~\cite{he2016_resnet}) generates separate multi-scale horizontal-view and vertical-view radar feature maps:
$\Zbf_{\mathtt{hor}}  = \{ \Zbf_{\mathtt{hor},i}\}_{i=1}^{S}=\mathtt{backbone}\left(\Ybf_{\mathtt{hor}}\right)$ and $\Zbf_{\mathtt{ver}} = \{ \Zbf_{\mathtt{ver},i}\}_{i=1}^{S}=\mathtt{backbone}\left(\Ybf_{\mathtt{ver}}\right)$, where the $i$-th scale feature maps $\Zbf_{\mathtt{hor},i}\in\mathbb{R}^{{W_i}\times D_i \times d}$ and $\Zbf_{\mathtt{ver},i}\in\mathbb{R}^{{H_i}\times D_i\times d}$ have a spatial dimension of $W_i \times D_i$ or  $H_i \times D_i$  and a feature dimension of $d$, and $S$ is the number of scales. 

\textbf{Cross-View Encoder}
is a Transformer encoder with $L_{\mathtt{enc}}$ layers that fuses the horizontal- and vertical-view radar features.
Each layer runs a shared cross-attention twice: first with $\Zbf_\mathtt{hor}$ as key/value and $\Zbf_\mathtt{ver}$ as query, then vice versa.
This bidirectional exchange embeds complementary cues, while residual connections keep view-specific details, producing refined features $\Fbf_\mathtt{enc}^{(i)}, \ i = 1, \cdots, L_\mathtt{enc}$,
\begin{equation}
\Fbf_{a}^{(i)} = \Fbf_{a}^{(i-1)} + \mathtt{CrossAttn}(\Fbf_{a}^{(i-1)}, \Fbf_{b}^{(i-1)}), \quad (a,b) \in \{(\mathtt{hor}, \mathtt{ver}), (\mathtt{ver}, \mathtt{hor})\},
\end{equation}
where $\Fbf_\mathtt{hor}^{(0)} = \Zbf_\mathtt{hor}$, $\Fbf_\mathtt{ver}^{(0)} = \Zbf_\mathtt{ver}$, 
and $\mathtt{CrossAttn}(\cdot, \cdot)$ denotes the deformable cross-attention~\cite{zhu2021_deformable_detr} following \cite{shi2022_petr} with fixed positional embeddings added beforehand for efficiency.  
After $L_{\mathtt{enc}}$ iterations, the encoded features $\Fbf_\mathtt{hor}$ and $\Fbf_\mathtt{ver}$ are obtained at the output of the cross-view encoder. 

\textbf{Pseudo-3D Pose Decoder} associates $N$ pose queries $\Qbf_\mathtt{pose} \in \mathbb{R}^{N\times d}$ (embedding dimension $d$) with encoded radar features $(\Fbf_\mathtt{hor}, \Fbf_\mathtt{ver})$, where each query corresponds to a reference pose refined through pseudo-3D deformable attention over $L_\mathtt{pose}$ layers. 
We define the $l$-th decoder layer as a function ${\mathcal{D}^{(l)}_{\mathtt{pose}}}$ that updates both the pose queries and reference poses in the 3D radar space:
\begin{equation}
( \Qbf_\mathtt{pose}^{(l)}, \tilde{\Pbf}^{(l)}_\mathtt{radar} ) = \mathcal{D}_{\mathtt{pose}}^{(l)}( \Qbf_\mathtt{pose}^{(l-1)}, \Fbf_\mathtt{hor}, \Fbf_\mathtt{ver}, \tilde{\Pbf}^{(l-1)}_\mathtt{radar} ),
\end{equation}
where $\tilde{\Pbf}^{(0)}_\mathtt{radar}$ is initialized by passing $\Qbf_\mathtt{pose}$ through an MLP. 
Reference poses are iteratively refined by applying predicted coordinate offsets $\Delta\tilde{\Pbf}^{(l)}_{\mathtt{radar}}$ in the normalized scale:
\begin{equation}
{\tilde{\Pbf}^{(l)}_\mathtt{radar} \in \mathbb{R}^{N\times 3K} = \sigma(\sigma^{-1}(\tilde{\Pbf}^{(l-1)}_\mathtt{radar}) + \Delta\tilde{\Pbf}^{(l-1)}_\mathtt{radar}), \quad l = 1, \ldots, L_\mathtt{pose}},
\end{equation}
where $\sigma$ and $\sigma^{-1}$ denote the Sigmoid function and its inverse. The predicted offsets ${\Delta \tilde{\Pbf}^{(l)}_{\mathtt{radar}}}  = { {H_{\mathtt{pose}}}(\Qbf_{\mathtt{pose}}^{(l)})}$ are obtained by passing pose queries at each layer to a shared regression head ${H_\mathtt{pose}}$. 

We convert the initial pose estimates $\tilde{\Pbf}_{\mathtt{radar}} = {\tilde{\Pbf}_\mathtt{radar}^{(L_\mathtt{pose})}}$ from the radar coordinate system to the world coordinate system via ${\tilde{\Pbf}_\mathtt{world}} = \mathcal{T}_{\mathtt{r2w}} ( \tilde{\Pbf}_\mathtt{radar} )$, along with the corresponding confidence scores $\tilde{\cbf}$. 
We defer the pseudo-3D deformable attention to Section~\ref{sec:pseudo_3D_deformable_attention}. 

\textbf{Pseudo-3D Joint Decoder} associates $K$ joint queries $\Qbf_\mathtt{joint} \in \mathbb{R}^{K\times d}$ with encoded radar features $(\Fbf_\mathtt{hor}, \Fbf_\mathtt{ver})$, where each query corresponds to a single joint refined by pseudo-3D deformable attention over $L_{\mathtt{joint}}$ layers.
Here, $K$ joint queries correspond to the same subject.
We define the $l$-th decoder layer as a function $\mathcal{D}^{(l)}_\mathtt{joint}$ that updates both the joint queries and corresponding joints:
\begin{align}
    ( \Qbf_\mathtt{joint}^{(l)}, \tilde{\pbf}^{(l)}_{i, \mathtt{radar}} ) = \mathcal{D}_{\mathtt{joint}}^{(l)}( \Qbf_\mathtt{joint}^{(l-1)}, \Fbf_\mathtt{hor}, \Fbf_\mathtt{ver}, \tilde{\pbf}^{(l-1)}_{i, \mathtt{radar}} ),
\end{align}
where $\tilde{\pbf}^{(l)}_{i, \mathtt{radar}} \in \mathbb{R}^{K\times 3}, i=1,\cdots,N$ is one specific pose in the $N$ poses, and $\tilde{\pbf}^{(0)}_{i, \mathtt{radar}}$ is $i$-th pose prediction from the pose decoder.
Joints in the reference pose are iteratively refined by applying predicted coordinate offsets ${\Delta\tilde{\pbf}_{i, {\mathtt{radar}}}^{(l)}}$:
\begin{align}
    \tilde{\pbf}^{(l)}_{i, \mathtt{radar}} \in \mathbb{R}^{K\times 3}= \sigma(\sigma^{-1}(\tilde{\pbf}^{(l-1)}_{i, \mathtt{radar}}) + \Delta\tilde{\pbf}^{(l-1)}_{i, \mathtt{radar}}), \quad l=1, \cdots, L_\mathtt{joint},
\end{align}
where the predicted offsets are given as $\Delta\tilde{\pbf}_{i, {\mathtt{radar}}}^{(l)} = H_\mathtt{joint}(\Qbf_{\mathtt{joint}}^{(l)})$ with a shared regression head. 

We collect refined reference poses from the joint decoder as $\hat{\Pbf}_{\mathtt{radar}} = \{\tilde{\pbf}_{i, {\mathtt{radar}}}^{(L_{\mathtt{joint})}}\}_{i=1}^{N}$ and convert them into the 3D world coordinate system as  ${\hat{\Pbf}_{\mathtt{world} }= {\mathcal{T}}_{\mathtt{r2w}}(\hat{\Pbf}_{\mathtt{radar}})}$. 

\subsection{Pseudo-3D Deformable Attention}
\label{sec:pseudo_3D_deformable_attention}
\begin{wrapfigure}[18]{r}{3.2in}
    \vspace{-0.0in}
    \centering
    \includegraphics[width=\linewidth]{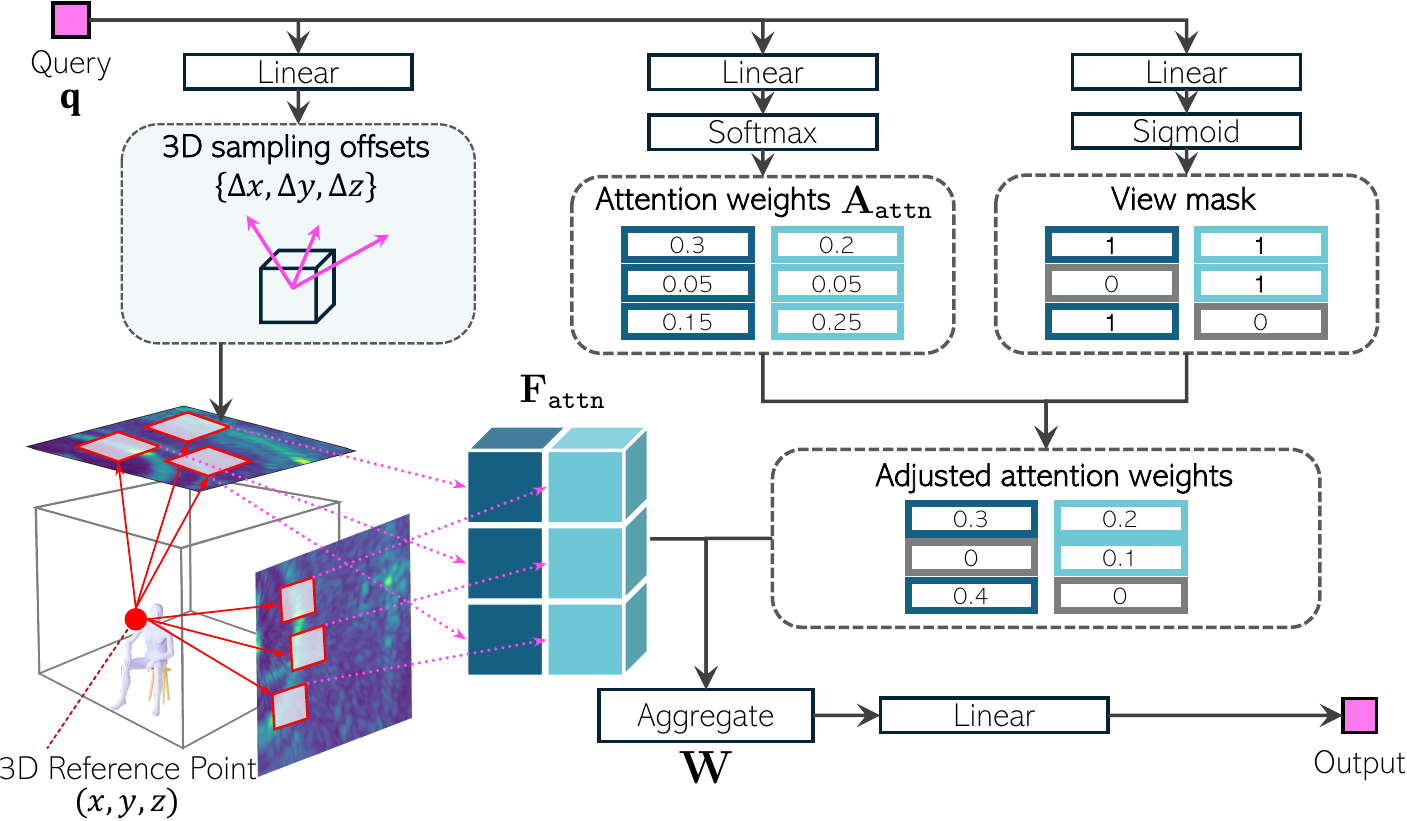}
    \caption{The pseudo-3D deformable attention operates on a 3D reference point and 3D sampling offsets that are projected to different radar views for pseudo-3D attention between multi-view radar features and the query.}
    \label{fig:deform_attn}
    \vspace{20mm}
\end{wrapfigure}
Our two-stage decoder incorporates a pseudo-3D deformable attention module, where ``pseudo'' highlights that reference points and sampling offsets are defined in 3D space, while feature sampling occurs on the 2D radar views, as illustrated in Fig.~\ref{fig:deform_attn}.

Consider a 3D reference point $(x, y, z)$ in the 3D radar space with a corresponding query $\qbf \in \Rset^{d}$ (from either pose queries in the pose decoder or joint queries in the joint decoder). 
We first feed $\qbf$ into a linear projection layer to predict a set of 3D sampling offsets $\{\Delta x_i, \Delta y_i, \Delta z_i\}_{i=1}^{N_\mathtt{offset}}$.  Given the 3D reference point and sampling offsets, we can locate the 3D sampling coordinates and project them onto the two radar views, extracting deformable multi-view radar features:
\begin{equation}\label{eq:feature_sampling}
    \fbf_\mathtt{hor}^{(i)} = \Fbf_\mathtt{hor}(x+\Delta x_i, z + \Delta z_i),  \quad
    \fbf_\mathtt{ver}^{(i)} = \Fbf_\mathtt{ver}(y+\Delta y_i, z + \Delta z_i) \quad i=1, \cdots, N_\mathtt{offset}.
\end{equation}
We group deformable multi-view radar features as $\Fbf_\mathtt{attn} = \{\fbf_{\mathtt{hor}}^{(1)}, \fbf_{\mathtt{ver}}^{(1)}, \cdots, \fbf_{\mathtt{hor}}^{(N_\mathtt{offset})}, \fbf_{\mathtt{ver}}^{(N_\mathtt{offset})}\}$.

Meanwhile, multi-view attention weights $f_\mathtt{attn} \in \mathbb{R}^{N_\mathtt{offset}\times 2}$ (where $2$ corresponds to the two radar views) are proposed by linearly projecting the query and applying a softmax normalization. These weights capture the relative importance of radar features across the two views in a unified manner.

Given the deformable multi-view radar features $\Fbf_\mathtt{attn}$ and the multi-view attention weights $f_\mathtt{attn}$, deformable multi-view attention features can be calculated as
\begin{equation}\label{eq:pseudo_3d_deformable_attention}
    \bar{\Fbf}_{\mathtt{attn}} = \sum^{N_{\mathtt{offset}}}_{i=1} (A_{i,0}  \Wbf \Fbf_\mathtt{attn}^{(2i-1)} + A_{i,1}  \Wbf \Fbf_\mathtt{attn}^{(2i)}),
\end{equation}
where $A_{i,0}$ and $A_{i,1}$ are the attention weights in $f_\mathtt{attn}$ for the $i$-th deformable radar feature in the horizontal and, respectively, vertical radar views,  and $\Wbf \in \mathbb{R}^{C\times C}$ is a learnable weight matrix.
We denote the overall pseudo-3D deformable attention as $\bar{\Fbf}_{\mathtt{attn}} = {\mathtt{DeformableAttn}}(\Fbf_{\mathtt{ver}}, \Fbf_{\mathtt{hor}}, (x, y, z), \qbf)$. 

Appendix~\ref{app:multi_head_attention} provides implementation details of ${\mathtt{DeformableAttn}}(\cdot, \cdot, \cdot, \cdot)$ and a computational complexity comparison with the decoupled 2D deformable attention used in QRFPose~\cite{wan2024qrfpose}, demonstrating better scalability of the proposed pseudo-3D attention as the number of radar views increases. 
Appendix~\ref{app:avs} describes an optional view mask module (top right of Fig.~\ref{fig:deform_attn}) that adds flexibility in selecting multi-view radar features per query. For example, an all-zero mask can be applied to exclude features from a specific radar view.

\subsection{Structural Loss Function}
\label{sec:loss_function}
As illustrated in Fig.~\ref{fig:architecture}, RAPTR utilizes weak supervision labels: coarse-grained BBox labels $\Bbf_{\mathtt{world}}$ in the 3D world coordinate system and fine-grained 2D keypoint labels $\Pbf_{\mathtt{image}}$ in the image plane. The loss function is calculated between these labels $\{\Bbf_{\mathtt{world}},  \Pbf_{\mathtt{image}}\}$ and the initial and refined 3D pose estimates $\{ \tilde{\Pbf}_{\mathtt{world} }, \hat{\Pbf}_{\mathtt{world} } \}$ with details included in Appendix~\ref{app:loss}.

\textbf{3D Template (T3D) Loss at Pose Decoder} utilizes coarse-grained 3D BBox labels $\Bbf_\mathtt{world}$. For each $\Bbf_\mathtt{world}$, we construct a 3D keypoint template by computing the centroid of the corresponding 3D BBox, which serves as the 3D gravity center label $\gbf_\mathtt{world} \in \Rset^{1 \times 3}$. 

Then, given a keypoint template defined at the coordinate origin $\Kbf_\mathtt{world} \in \Rset^{K \times 3}$, the corresponding template pose $\Tbf_\mathtt{world}$ is computed as $\Tbf_\mathtt{world} = \Kbf_\mathtt{world} + \mathbf{1}^{\top} \gbf_\mathtt{world}$.
As illustrated in the lower right of Fig.~\ref{fig:architecture}, the T3D loss $\mathcal{L}_\mathtt{template}$ is defined as the Euclidean distance between the template poses $\Tbf_\mathtt{world}$ and the initial 3D pose estimates $\tilde{\Pbf}_\mathtt{world}$ at the pose decoder. 

\textbf{Combined 3D Gravity (G3D) Loss and 2D Keypoint (K2D) Loss at Joint Decoder} utilizes both the coarse-grained 3D BBox labels $\Bbf_\mathtt{world}$ and the fine-grained 2D keypoint labels $\Pbf_{\mathtt{image}}$ in the image plane, as illustrated in the upper right of Fig.~\ref{fig:architecture}

For the G3D loss, the refined 3D pose estimate $\hat{\Pbf}_\mathtt{world}$ is collapsed into its centroid as $\hat{\gbf}_\mathtt{world} \in \Rset^{1 \times 3}$ by averaging the keypoint coordinates along each spatial axis. The resulting G3D loss $\mathcal{L}_\mathtt{gravity}$ is then defined as the Euclidean distance between the predicted and ground-truth 3D gravity centers, $\hat{\gbf}_\mathtt{world}$ and $\gbf_\mathtt{world}$.

For the K2D loss, the refined 3D pose estimate $\hat{\Pbf}_{\mathtt{radar}}$ in the radar coordinate system are first transformed into the 3D camera coordinate system via a calibrated coordinate transformation: $\hat{\Pbf}_{\mathtt{camera}} = \Rbf\hat{\Pbf}_{\mathtt{radar}} + \mathbf{1}^{\top}\tbf$ where $\Rbf$ and $\tbf$ denote the calibrated 3D rotation matrix and the translation vector, respectively. 
The resulting 3D camera-space pose estimates are then projected onto the 2D image plane via a known 3D-to-2D projection: $\hat{\Pbf}_{\mathtt{image}} =\mathtt{proj}_{\mathtt{image}}(\hat{\Pbf}_{\mathtt{camera}})$.
Finally, the fine-grained 2D loss combines the image-plane Euclidean error $\mathcal{L}_\mathtt{kpt2D}$ and the object keypoint similarity (OKS) loss $\mathcal{L}_\mathtt{OKS}$~\cite{shi2022_petr} between $\Pbf_\mathtt{image}$ and $\hat{\Pbf}_\mathtt{image}$.

\textbf{Structural Loss Function}: 
Following the set-based loss in \cite{carion2020detr}, we employ bipartite matching to associate predictions $\{ {\hat{\gbf}_\mathtt{world}, \tilde{\Pbf}_\mathtt{world}, \hat{\Pbf}_\mathtt{world} } \}$ with their ground-truth labels $\{ {\gbf_\mathtt{world}, \Tbf_\mathtt{world}, \Pbf_\mathtt{world} } \}$. Based on these associations, we define the structural loss function as
\begin{align}
\mathcal{L} = \frac{1}{N^\prime} \sum_{i=1}^{N^\prime} (\lambda_1\mathcal{L}_\mathtt{template} + \lambda_2\mathcal{L}_\mathtt{gravity} + \lambda_3\mathcal{L}_\mathtt{kpt2D} + \lambda_4\mathcal{L}_\mathtt{OKS})+ \lambda_5\mathcal{L}_{\mathtt{cls}},
\end{align}
where $N^\prime$ is the number of matched pairs, $\lambda_i$ is the corresponding weighting factor for each loss term, and $\mathcal{L}_\mathtt{cls}$ is the classification loss of the focal loss~\cite{lin2020focal} with the confidence scores of the matched estimates.

\section{Evaluation}
\label{sec:evaluation}

\subsection{Settings}
\paragraph{Datasets:}
We assess the performance of RAPTR and baseline models on the HIBER dataset\footnote{\url{https://github.com/Intelligent-Perception-Lab/HIBER}}~\cite{RFMask23} and the MMVR dataset\footnote{\url{https://zenodo.org/records/12611978}}~\cite{Rahman2024_mmvr}, both of which are publicly available multi-view mmWave radar datasets designed for indoor human perception tasks.
The HIBER dataset includes two-view radar heatmaps from 10 different viewpoints, the corresponding 3D keypoint labels, and the 3D BBox labels.
We use data protocols ``MULTI'' and ``WALK'', and use views 2 through 10 for training, validation, and testing.
The MMVR dataset includes two-view radar heatmaps in various indoor scenarios, the corresponding 2D keypoint labels, and the 3D BBoxes.
We use a data split ``P1S1'', a single-person case in an open space.
A detailed description of the datasets is provided in Appendix~\ref{app:dataset}.

\paragraph{Parameter Settings for RAPTR:}
We use $T=4$ consecutive frames as input to our RAPTR network.
For the point decoder, the number of pose queries $N$ is 10.
For the joint decoder, the number of joint queries $K$ depends on the dataset to be evaluated: $K=14$ for HIBER and $K=17$ for MMVR.
The parameters relating to model training are summarized in Appendix~\ref{app:training_parameter}.

\paragraph{Baselines:}
We consider the following competitive radar/RF-based 3D pose estimation baselines:  \textbf{Person-in-WiFi 3D}~\cite{yan2024person}, \textbf{HRRadarPose}~\cite{yuan2024rtpose}, and \textbf{QRFPose}~\cite{wan2024qrfpose}.
We evaluate Person-in-WiFi 3D and HRRadarPose using their open-source implementations. As QRFPose has no public code, we reimplement it from scratches and verify similar performance to the original report~\cite{wan2024qrfpose} using 3D keypoint labels. 
For fair comparison, we adopt a loss function, similar to RAPTR, combining 2D keypoint loss and 3D gravity loss.
Baseline implementation details are provided in Appendix~\ref{app:baseline}.

\paragraph{Metrics:}
We employ \textbf{Mean Per Joint Pose Error~(MPJPE)} with the unit of centimeters in the world coordinate.
In addition, we evaluate this MPJPE for each body joint and along each 3D axis, horizontal (h), vertical (v), and depth (d), independently.
For MMVR, since 3D keypoint labels are not available, we construct a 3D \ac{BBox} that encloses the estimated 3D keypoints and then use \textbf{the distance between the center points} of this \ac{BBox} and the 3D \ac{BBox} labels, as well as \textbf{the absolute error in the lengths of the edges} along each axis of the box, as metrics to approximate the 3D pose estimation performance.
Detailed evaluation metrics are described in Appendix~\ref{app:metrics}.

\subsection{Main results}
\begin{figure}[tbp]
    \centering
    \begin{minipage}[b]{0.925\hsize}
        \centering
        \includegraphics[width=\linewidth]{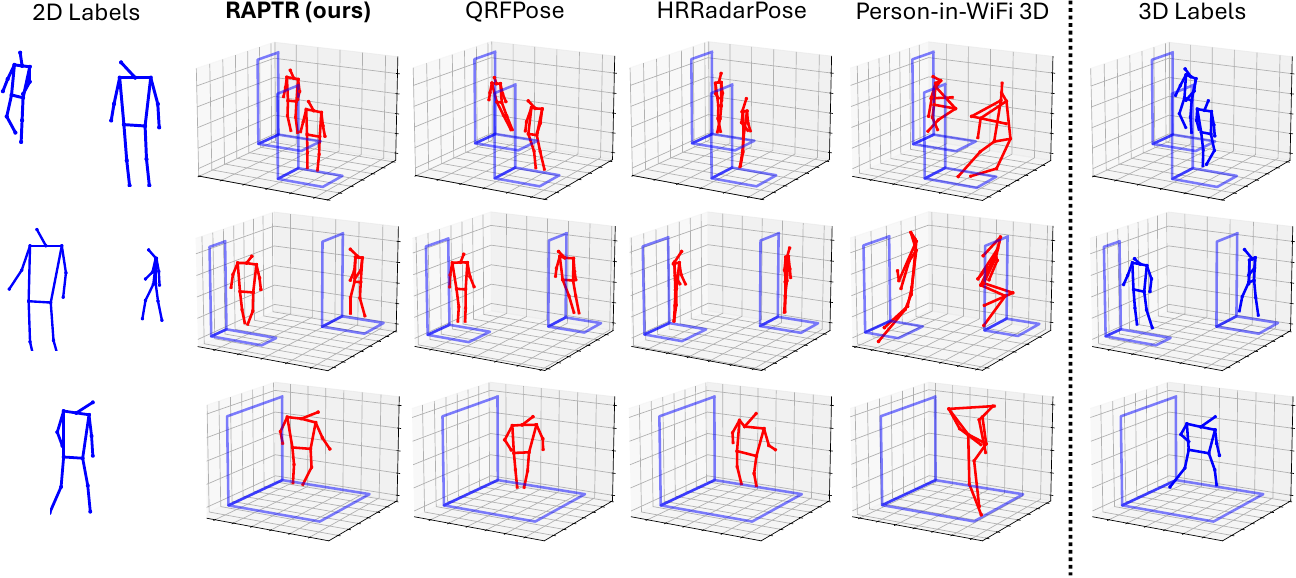}
        \subcaption{Visualization of 3D pose estimation by RAPTR and baseline methods on the HIBER dataset.}
        \vspace{5mm}
        \label{fig:vis_hiber_main}
    \end{minipage}
    \begin{minipage}[b]{0.85\hsize}
        \centering
        \includegraphics[width=0.85\linewidth]{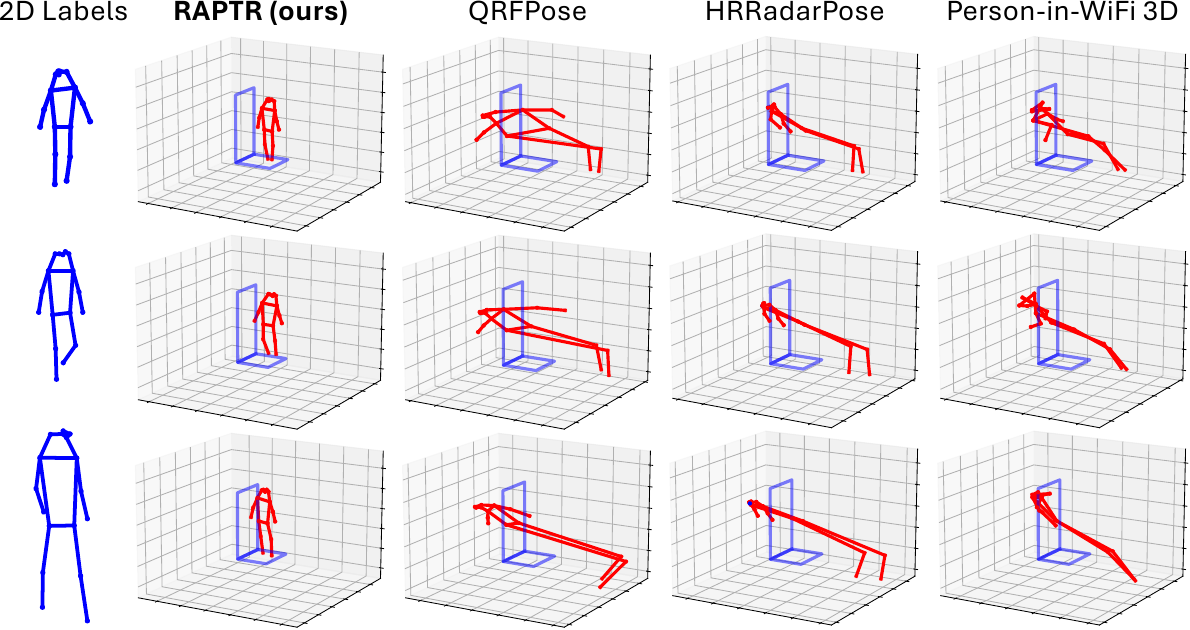}
        \subcaption{Visualization of 3D pose estimation by RAPTR and baseline methods on the MMVR dataset.}
        \label{fig:vis_mmvr_main}
    \end{minipage}
    \caption{Qualitative results. Blue lines indicate the keypoint labels, blue rectangles indicate the 3D BBox labels, and red lines indicate the predictions.}
    \vspace{-3mm}
\end{figure}

\paragraph{HIBER:}
Table~\ref{tab:main_result_hiber} shows the performance of 3D pose estimation for HIBER, using 2D and coarse 3D labels for baselines and our RAPTR.
The qualitative results are provided in Fig.~\ref{fig:vis_hiber_main}.

For WALK, RAPTR achieves a significantly lower overall MPJPE of \SI{22.32}{\centi\meter} and outperforms all other baselines in the metric.
More specifically, RAPTR reduces the overall error by 61.7\%, 41.6\%, and 34.3\% compared to Person-in-WiFi 3D, QRFPose, and HRRadarPose, respectively.
The per-joint breakdown demonstrates that RAPTR maintains its performance on relatively challenging joints, such as the wrist and ankle, where other baselines exhibit significant degradation.
For example, HRRadarPose reports a wrist error of \SI{42.33}{\centi\meter}, whereas RAPTR reports an error of \SI{26.55}{\centi\meter}.
Moreover, RAPTR maintains the error gap between the best- and worst-estimated joints within \SI{10}{\centi\meter}, showing a consistent level of accuracy throughout the body.
In terms of directional components, RAPTR shows much lower errors in the horizontal and vertical dimensions than baselines, indicating that RAPTR estimates well-proportioned 3D poses across all axes.

For MULTI, the more challenging multi-person scenario, RAPTR continues to outperform with an overall MPJPE of \SI{18.99}{\centi\meter} and shows a substantial margin compared to the second-best HRRadarPose at \SI{33.19}{\centi\meter}.
RAPTR reduces the overall error by 77.7\%, 58.8\%, and 42.7\% compared to Person-in-WiFi 3D, QRFPose, and HRRadarPose, respectively.
Although the overall accuracy of Person-in-WiFi 3D and QRFPose, noticeably degrades on the MULTI split compared to WALK, likely due to the increased complexity of handling multiple objects, RAPTR maintains a nearly consistent level of performance.

Referring to the qualitative results provided in Fig.~\ref{fig:vis_hiber_main}, RAPTR estimates structurally consistent 3D poses that match the 3D labels in both position and orientation, while baselines often suffer from misaligned limbs and implausible joint configurations.
While baselines often fail to maintain human-like pose structure in the MULTI setting despite performing well in WALK, RAPTR consistently produces plausible estimates in both scenarios, indicating its robustness to multi-person scenes.

\begin{table}[t]
    \scriptsize
    \centering
    \setlength\tabcolsep{4.2pt}
    \vspace{-3mm}
    \caption{3D pose estimation performance on HIBER (MPJPE: \si{\centi\meter}).}
    \label{tab:main_result_hiber}
    \begin{tabular}{c|c|cccccccc|cccc} \toprule
         \textbf{Env} & \textbf{Method} & \textbf{Head} & \textbf{Neck} & \textbf{Shoulder} & \textbf{Elbow} & \textbf{Wrist} & \textbf{Hip} & \textbf{Knee} & \textbf{Ankle} & \textbf{Overall} & \textbf{(h)} & \textbf{(v)} & \textbf{(d)} \\ \midrule
         \multirow{4}{*}{\rotatebox{90}{WALK}}& Person-in-WiFi 3D & 54.28 & 57.01 & 54.18 & 54.81 & 59.98 & 53.98 & 60.32 & 68.84 & 58.25 & 25.60 & 23.94 & 36.20 \\
         & QRFPose & 42.23 & 34.21 & 37.37 & 38.05 & 41.25 & 31.24 & 34.39 & 46.87 & 38.20 & 14.78 & 13.40 & 26.76 \\
         & HRRadarPose & 30.23 & 25.44 & 33.70 & 34.15 & 42.33 & 27.71 & 31.55 & 40.46 & 33.96 & 15.14 & 13.13 & 19.85 \\
         & {RAPTR (ours)} & \cellcolor{gray!20}21.75 & \cellcolor{gray!20}17.41 & \cellcolor{gray!20}20.72 & \cellcolor{gray!20}23.23 & \cellcolor{gray!20}26.55 & \cellcolor{gray!20}18.97 & \cellcolor{gray!20}21.06 & \cellcolor{gray!20}26.10 & \cellcolor{gray!20}\textbf{22.32} & \cellcolor{gray!20}\textbf{8.41} & \cellcolor{gray!20}\textbf{4.85} & \cellcolor{gray!20}\textbf{17.73} \\ \midrule
         \multirow{4}{*}{\rotatebox{90}{MULTI}} & Person-in-WiFi 3D & 88.48 & 85.14 & 89.44 & 84.33 & 84.29 & 88.69 & 81.70 & 81.53 & 85.25 & 34.06 & 28.57 & 58.93 \\
         & QRFPose & 49.49 & 44.48 & 45.54 & 46.77 & 49.06 & 40.99 & 41.87 & 51.57 & 46.11 & 18.20 & 14.13 & 34.39 \\
         & HRRadarPose & 30.24 & 24.24 & 30.14 & 35.17 & 44.34 & 28.76 & 31.38 & 35.31 & 33.19 & 16.77 & 10.75 & 21.84 \\
         & {RAPTR (ours)} & \cellcolor{gray!20}18.39 & \cellcolor{gray!20}13.13 & \cellcolor{gray!20}16.44 & \cellcolor{gray!20}20.12 & \cellcolor{gray!20}24.62 & \cellcolor{gray!20}15.01 & \cellcolor{gray!20}17.76 & \cellcolor{gray!20}23.22 & \cellcolor{gray!20}\textbf{18.99} & \cellcolor{gray!20}\textbf{7.80} & \cellcolor{gray!20}\textbf{4.38} & \cellcolor{gray!20}\textbf{14.54} \\ \bottomrule
    \end{tabular}
    \vspace{-5mm}
\end{table}

\begin{table}[ht]
    \centering
    \scriptsize
    \setlength\tabcolsep{3.3pt}
    \vspace{-5mm}
    \caption{Pose estimation performance on MMVR (P1S1).}
    \label{tab:main_result_mmvr}
    \begin{tabular}{c|cccc} \toprule
         \multirow{2}{*}{\textbf{Method}} & \multirow{2}{*}{\textbf{Center distance (\si{\centi\meter})}} & \multicolumn{3}{c}{\textbf{Edge length error (\si{\centi\meter})}} \\
         \cline{3-5}
         & & (h) & (v) & (d) \\ \midrule
        Person-in-WiFi 3D & 136.14 & 33.18 & 95.43 & 242.86 \\
        QRFPose & 210.75 & 38.12 & 73.69 & 409.38 \\
        HRRadarPose & 164.46 & 37.84 & 74.00 & 313.81 \\ \midrule
        RAPTR (ours) & \cellcolor{gray!20}\textbf{31.41} & \cellcolor{gray!20}\textbf{22.90} & \cellcolor{gray!20}\textbf{10.66} & \cellcolor{gray!20}\textbf{50.56} \\ \bottomrule
    \end{tabular}
    \vspace{-5mm}
\end{table}

\paragraph{MMVR:}
Table~\ref{tab:main_result_mmvr} shows the performance comparison for baselines and RAPTR with MMVR, and Fig.~\ref{fig:vis_mmvr_main} provides qualitative results.
Although we cannot directly evaluate the precise 3D pose estimation performance for MMVR due to the absence of 3D pose labels, the results demonstrate that RAPTR effectively preserves reasonable human pose and location accuracy in the 3D space.
Specifically, RAPTR shows improvements in center distance by 76.9\%, 85.1\%, and 80.9\% compared to Person-in-WiFi 3D, QRFPose, and HRRadarPose, respectively.
As shown in Fig.~\ref{fig:vis_mmvr_main}, other baselines exhibit degraded performance due to structural collapse in 3D space, caused by overfitting to 2D alignment when projected onto the image plane. 
We assume that RAPTR effectively avoids this issue by not directly predicting the keypoints, but instead refining the final output through 2D keypoint supervision applied to each joint of a template pose that is placed in the 3D space.

\section{Ablation Study}
\label{sec:ablation}

In this section, we present ablation studies of our RAPTR on the HIBER dataset. Unless otherwise stated, all reported evaluation results are reported as the mean $\pm$ standard deviation, computed over three random seeds. Additional ablation results and visualizations are provided in Appendix~\ref{app:ablation}.

\paragraph{Visualization of Pose Refinement Process:}
\begin{figure}[t]
    \centering
    \includegraphics[width=0.8\linewidth]{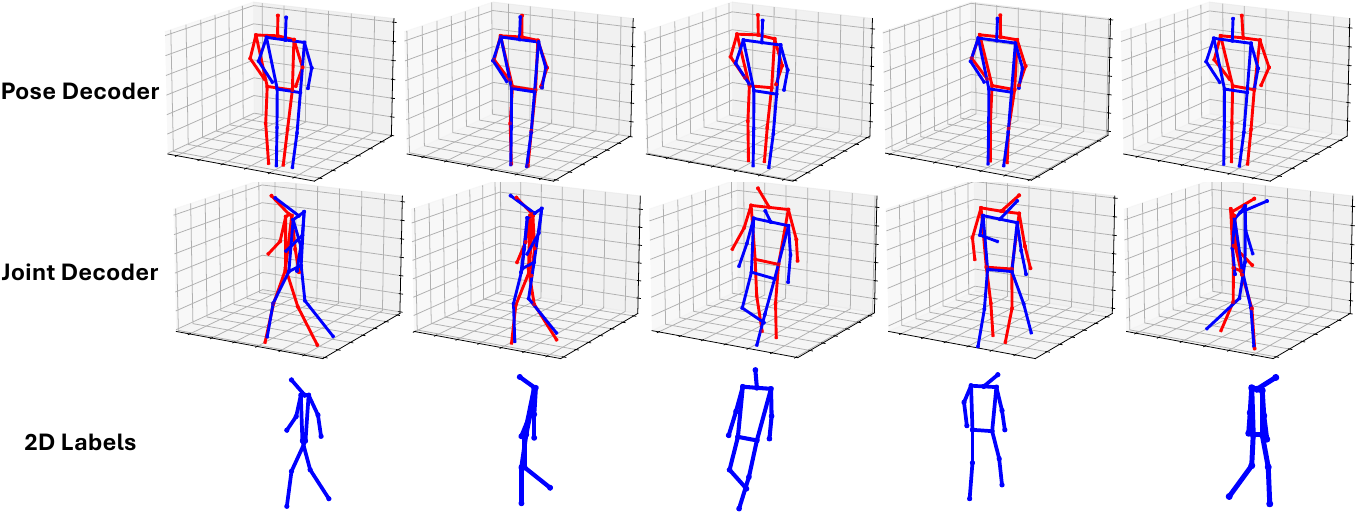}
    \caption{Pose Refinement Process: the pose prediction is first constrained by the 3D template at the pose decoder, and subsequently refined at the joint decoder.}
    \label{fig:pose_refine_process}
    \vspace{-0mm}
\end{figure}
Fig.~\ref{fig:pose_refine_process} illustrates the refinement process of   a 3D prediction through the two-stage decoder architecture.
The pose decoder first establishes coarse 3D structures of the human body under the constraint of the 3D template loss (1st row). 
Subsequently, the joint decoder fine-tunes the keypoints to better capture the subject orientation and limb configuration (2nd row), while preserving the structure consistency provided by the pose decoder.

\paragraph{Effect of Loss Terms:} Table~\ref{tab:ablation_loss_function} provides an ablation study on the effect of different combinations of loss terms in the two-stage decoder. When only the K2D loss is applied at the joint decoder (row $1$), the 3D pose estimation suffers from depth ambiguity due to the absence of any 3D constraint, resulting in a substantial increase in MPJPE to \SI{381.18}{\centi\meter} and \SI{375.73}{\centi\meter} on the WALK and MULTI splits, respectively. From rows~2 to~4 in Table~\ref{tab:ablation_loss_function}, we remove or modify one loss term at a time from the proposed structural loss.
Removing the T3D loss at the pose decoder (row $2$), replacing the T3D with the K2D+G3D loss at the pose decoder (row $3$), or removing the G3D loss at the joint decoder (row $4$) leads to a noticeable degradation in 3D pose estimation.

\begin{threeparttable}[t]
    \centering
    \footnotesize
    \setlength\tabcolsep{3pt}
    \caption{Effect of loss terms for RAPTR (MPJPE: \si{\centi\meter}).}
    \begin{tabular}{cccc|l} \toprule
            \multicolumn{2}{c}{\textbf{Loss}} & \multirow{2}{*}{\textbf{WALK}} & \multirow{2}{*}{\textbf{MULTI}} & \multirow{2}{*}{\textbf{Notes}} \\
            \cline{1-2}
            Pose Dec. & Joint Dec. & & & \\
            \midrule
            -- & \texttt{K2D} & 381.18 $\pm$ 0.28 & 375.73 $\pm$ 6.31 &  2D keypoint loss only at joint decoder \\
            \midrule
            -- & \texttt{K2D+G3D} & 28.54 $\pm$ 4.57 & 57.90 $\pm$ 9.81 & without 3D template loss at pose decoder  \\
            \texttt{K2D+G3D} & \texttt{K2D+G3D} & 27.49 $\pm$ 3.40 & 23.43 $\pm$ 3.44 & with 2D keypoint + 3D gravity loss at both decoders \\
            \texttt{T3D} & \texttt{K2D} & 25.96 $\pm$ 4.95 & 25.83 $\pm$ 3.87 & without 3D gravity loss at joint decoder \\
            \midrule
            \texttt{T3D} & \texttt{K2D+G3D} & \textbf{22.32 $\pm$ 0.06} & \textbf{18.99 $\pm$ 0.16} & proposed structural loss \\
         \bottomrule
    \end{tabular}
    \label{tab:ablation_loss_function}
    \begin{tablenotes}[para,flushleft,online,normal]
        \item[] \texttt{K2D} = 2D Keypoint loss, \texttt{T3D} = 3D Template loss, \texttt{G3D} = 3D Gravity loss
    \end{tablenotes}
\end{threeparttable}

\paragraph{Effect of Deformable Attention Mechanisms:} Table~\ref{tab:ablation_pseudo_3d} presents an ablation study on the effect of the deformable attention mechanism for RAPTR. In this study, the pseudo-3D deformable attention is replaced with the decoupled 2D deformable attention used in QRFPose~\cite{wan2024qrfpose}, while keeping the cross-view encoder, two-stage decoder architecture, and the proposed structural loss unchanged. The results show that the pseudo-3D deformable attention yields marginal performance improvements, approximately $4$\% and $2.5$\% on the WALK and MULTI splits, respectively.
\begin{table}[htb]
    \centering
    \footnotesize
    \setlength\tabcolsep{3pt}
    \caption{Effect of deformable attention mechanisms for RAPTR (MPJPE: \si{\centi\meter}).}
    \begin{tabular}{ccc|l} \toprule
            \textbf{Attn.} & \textbf{WALK} & \textbf{MULTI} & \textbf{Notes} \\
            \midrule
            \texttt{2D} & 23.25 $\pm$ 1.38 & 19.47 $\pm$ 0.95 & RAPTR with decoupled 2D deformable attention \\
            \texttt{3D} & \textbf{22.32 $\pm$ 0.06} & \textbf{18.99 $\pm$ 0.16} & RAPTR with pseudo-3D deformable attention \\
         \bottomrule
    \end{tabular}
    \label{tab:ablation_pseudo_3d}
\end{table}

\paragraph{Comparison with a 2D-to-3D Pose Uplifting Model:}
\begin{table}[t]
    \centering
    \footnotesize
    \setlength\tabcolsep{3pt}
    \caption{Comparison with a 2D-to-3D pose uplifting model (MPJPE: \si{\centi\meter}).}
    \begin{tabular}{ccccc} \toprule
        & \multicolumn{2}{c}{\textbf{Loss}} & \multirow{2}{*}{\textbf{WALK}} & \multirow{2}{*}{\textbf{MULTI}}\\
        \cline{2-3}
        & Pose Dec. & Joint Dec. & & \\
        \midrule
        Pose Lifting~\cite{martinez2017simple} & \texttt{K2D} & \texttt{K2D} & 43.43 $\pm$ 2.66 & 41.76 $\pm$ 6.85 \\
        RAPTR (ours) & \texttt{T3D} & \texttt{K2D+G3D} & \textbf{22.32 $\pm$ 0.06} & \textbf{18.99 $\pm$ 0.16} \\
        \bottomrule
    \end{tabular}
    \label{tab:ablation_uplifting}
\end{table}
We further compare RAPTR with a baseline that first estimates 2D keypoints in the image plane and subsequently lifts them to 3D space using a pre-trained 2D-to-3D pose uplifting model~\cite{martinez2017simple} trained on vision-based datasets such as Human3.6M~\cite{ionescu2014human}. To ensure a fair comparison, this baseline adopts the same network architecture as RAPTR, but both the pose and joint decoders are supervised only by the 2D keypoint loss. Because the 3D poses predicted by the uplifting model are defined in a pelvis-centered coordinate system, we additionally estimate a translation offset to align the estimated poses with their correct position in the world coordinate system.
As shown in Table~\ref{tab:ablation_uplifting}, the pose uplifting baseline performs significantly worse than RAPTR, with MPJPEs of \SI{43.43}{\centi\meter} and \SI{41.76}{\centi\meter} on the WALK and MULTI splits, respectively.

\paragraph{Limitation:}
\label{sec:limitation}
Given that the process of refining the template to the actual pose in the joint decoder is supervised by the 2D keypoint labels, the accuracy of the 3D pose estimation is highly dependent on the precision of the labels in the image plane.
In this context, since the 2D keypoint label lacks the ability to discern whether the person is facing forward or backward to the camera, the estimated 3D poses may have joints that are bent in the opposite direction in depth from the actual pose.
In addition, real-world conditions such as occlusion and human-to-human interference can further degrade the pose estimation performance. These effects become more pronounced in crowded or interactive environments.

\section{Conclusion}
\label{sec:conclusion}
We introduced RAPTR, a radar-based 3D human pose estimation system using reliable 2D keypoint labels and 3D BBoxes as the coarse-grained 3D information.
We designed the network architecture and the loss function to integrate multi-view radar features and consistently represent human poses in the 3D space, whose effectiveness was demonstrated through experimental results.

\paragraph{Broader Impacts:}
Indoor radar perception technologies, such as RAPTR, provide diverse indoor applications.
These technologies may improve the safety and energy efficiency of indoor systems while preserving privacy. However, it is paramount that the perception results remain secure and private to prevent misuse.

{
    \small
    \bibliographystyle{ieeenat_fullname}
    \bibliography{neurips2025_raptr}
}

\clearpage
\appendix

\section{RAPTR Architecture}

\paragraph{Cross-view Encoder:}
\begin{figure}[b]
    \centering
    \begin{minipage}[b]{0.4\hsize}
        \centering
        \includegraphics[width=0.9\linewidth]{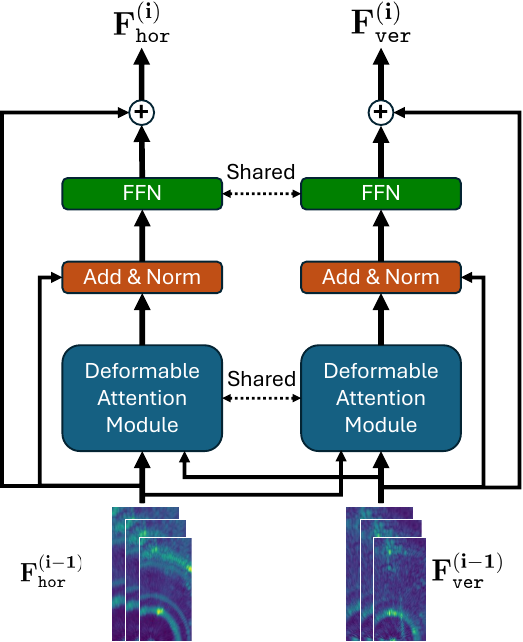}
        \subcaption{Cross-view encoder}
        \label{fig:app_architecture_encoder}
    \end{minipage}
    \begin{minipage}[b]{0.55\hsize}
        \centering
        \includegraphics[width=\linewidth]{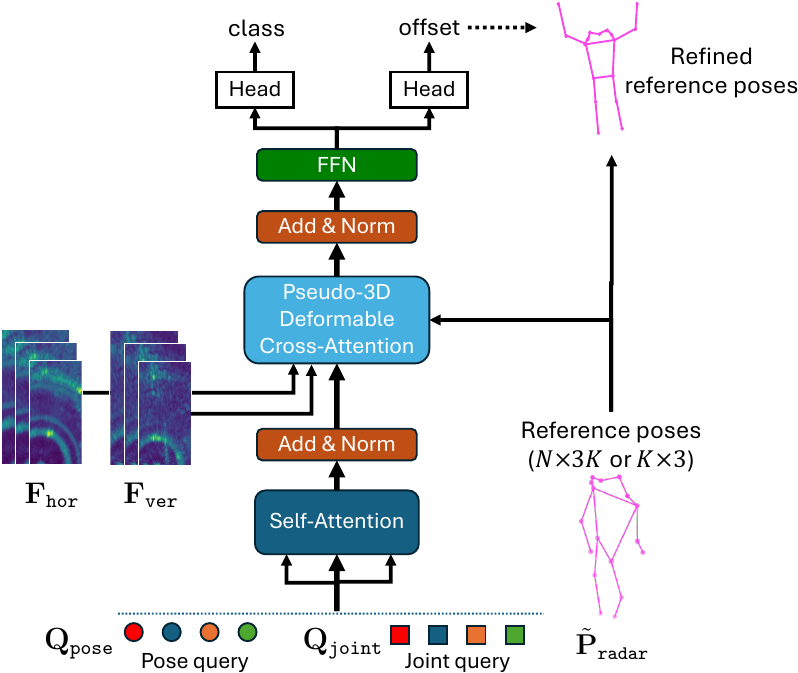}
        \subcaption{Pose/Joint decoder}
        \label{fig:app_architecture_decoder}
    \end{minipage}
    \caption{Transformer design in RAPTR.}
    \label{fig:app_architecture}
\end{figure}
Fig.~\ref{fig:app_architecture_encoder} illustrates the layer structure of the cross-view encoder in our RAPTR architecture.
The cross-view encoder associates multi-view radar features using the cross-attention and skip-connection mechanism.
A shared-weight backbone extracts sets of multi-scale features for both horizontal and vertical radar heatmaps as $\Zbf_\mathtt{hor} = \{\Zbf_{\mathtt{hor}, i}\}_{i=1}^{S}$ and $\Zbf_\mathtt{ver} = \{\Zbf_{\mathtt{ver}, i}\}_{i=1}^{S}$ with $S$ scale levels.
Each scale-level feature map is enriched with spatial positional encoding and a learnable level embedding, following \cite{zhu2021_deformable_detr}, and then fed into the cross-view encoder.
The cross-view encoder consists of a stack of $L_\mathtt{enc}$ multi-head multi-scale deformable attention layers, and we denote the input horizontal/vertical features to the $i$-th layer as $\Fbf_\mathtt{hor}^{(i)}, \Fbf_\mathtt{ver}^{(i)}$, respectively.
In our case, $\Fbf_\mathtt{hor}^{(0)} = \Zbf_\mathtt{hor}, \Fbf_\mathtt{ver}^{(0)} = \Zbf_\mathtt{ver}$.
Each $i$-th layer runs a shared cross-attention bidirectionally: first with $\Fbf_\mathtt{hor}^{(i-1)}$ as key/value and $\Fbf_\mathtt{ver}^{(i-1)}$ as query, then vice versa.
Residual connection, layer normalization, and an FFN follow as in standard Transformer encoders to further refine the features, and additional residual connections are incorporated to preserve view-specific details.
As a whole, the encoding process in the $i$-th layer is written as:

\begin{align}
\bar{\Fbf}_\mathtt{a}^{(i-1)} &= \Fbf_\mathtt{a}^{(i-1)} + \mathtt{CrossAttn}(\Fbf_\mathtt{a}^{(i-1)}, \Fbf_\mathtt{b}^{(i-1)}), \nonumber \\
\Fbf_\mathtt{a}^{(i)} &= \Fbf_\mathtt{a}^{(i-1)} + \mathtt{FFN}(\mathtt{layernorm}(\bar{\Fbf}_\mathtt{a}^{(i-1)})), \quad (a,b) \in \{(\mathtt{hor}, \mathtt{ver}),(\mathtt{hor}, \mathtt{ver})\}.
\end{align}

The output of the last layer, or encoder memory, is denoted as $\Fbf_\mathtt{hor}, \Fbf_\mathtt{ver}$.

\paragraph{Pose/Joint Decoder:}
Fig.~\ref{fig:app_architecture_decoder} illustrates the layer structure of the pose/joint decoder in RAPTR.
The pose decoder and the joint decoder share the architecture: they receive pose/joint queries $\Qbf_\mathtt{pose}, \Qbf_\mathtt{joint}$, multi-scale encoder memory from the cross-view encoder $\Fbf_\mathtt{hor}, \Fbf_\mathtt{ver}$, and reference points $\tilde{\Pbf}_\mathtt{radar}$.
The decoders then generate refined embeddings through multi-head self-attention and pseudo-3D deformable attention layers, which is the process denoted as $\mathcal{D}_\mathtt{pose}$ and $\mathcal{D}_\mathtt{joint}$ in Section~\ref{sec:methodology}.
$N$ pose queries correspond to $N$ pose predictions in the pose decoder, whereas $K$ joint queries correspond to $K$ joints on the same subject in the joint decoder.
The queries are first fed into self-attention, followed by residual connection and layer normalization, and then passed into the pseudo-3D deformable attention layer, as defined in Section~\ref{sec:pseudo_3D_deformable_attention}.
The pseudo-3D deformable attention layer is a cross-attention layer, using encoder memory to produce keys and values, which correlate with the refined queries.
In addition, reference points are fed into this layer to determine the sampling locations and aggregate sparse features on the multi-view encoder memory across space and scales in the pseudo-3D deformable attention mechanism.
The outputs are then passed through another residual connection, layer normalization, and an FFN.
In the pose decoder, a class regression head takes the resulting $N$ queries to calculate confidence scores $\hat{\cbf}$ for each corresponding person.
A pose regression head also takes the resulting queries to calculate pose coordinate offsets $\mathbb{R}^{N\times 3K}$ to refine the reference poses.
In the joint decoder, a pose regression head, shared with the pose decoder, takes the resulting $K$ queries to calculate the joint coordinate offsets $\mathbb{R}^{K\times 3}$ to refine the reference points.
The pose decoder and the joint decoder consist of $L_\mathtt{pose}, L_\mathtt{joint}$ decoder layers, respectively, and their outputs are the initial pose estimates and the refined pose estimates, as shown in Fig.~\ref{fig:architecture}.

In the implementation, we have a technically involved step regarding the application of bipartite matching to the batched estimations from the pose decoder.
We train the model using mini-batches, where the reference poses are first shaped as $B \times N \times 3K$, with $B$ denoting the mini-batch size.
While Section~\ref{sec:loss_function} describes the bipartite matching in general terms, it is in practice applied to the initial pose estimations $\tilde{\Pbf}_\mathtt{world}$ output by the pose decoder and the corresponding 3D keypoint labels $\Pbf_\mathtt{world}$ so that the computational cost of the subsequent joint decoder would be reduced.
The matching is guided by the regressed confidence scores $\hat{\cbf}$ and the Euclidean distance between the initial estimations and the labels.
Out of all estimations in the mini-batch, we retain only the $N^\prime$ matched ones and reshape them to $N^\prime \times K \times 3$ to serve as the reference poses for the joint decoder, in which $N^\prime$ is considered the new mini-batch size.

\paragraph{Computational Complexity:}
\label{app:network_architecture}
The cross-view encoder takes two sets of multi-scale feature maps $\Fbf_\mathtt{hor}, \Fbf_\mathtt{ver}$ for horizontal-depth and vertical-depth radar perceptions.
As shown in Fig.~\ref{fig:app_architecture_encoder}, we apply deformable cross-attention in both directions: from $\Fbf_\mathtt{hor}$ to $\Fbf_\mathtt{ver}$ and vice versa, treating one set as queries and the other as keys and values in each direction.
For each direction, given $S$-level feature scales, each with $N_s$ spatial positions, and $N_\mathtt{offset}$ sampling points per head, the total computational cost is
\begin{align}
    \mathcal{O}(2(d^2 + N_\mathtt{offset}d) \sum_{s=1}^S N_s) \approx \mathcal{O}((d^2 + N_\mathtt{offset}d) \sum_{s=1}^S N_s),
\end{align}
where $d$ is the feature dimension.

The pose decoder takes $N$ object queries and the encoded memory and performs self-attention and pseudo-3D deformable cross-attention.
Therefore, the computational cost is written as
\begin{align}
    \mathcal{O}(N^2 d + Nd^2 + N N_\mathtt{offset} S d).
\end{align}
Here, we omit the constant factor associated with bilinear interpolation and regression of sampling offsets and attention weight matrices in the last term, as it does not affect the asymptotic complexity.

Finally, the joint decoder takes $K$ joint queries for $N^\prime$ poses, selected through a bipartite matching procedure out of $N$, and the encoded memory and performs pseudo-3D deformable attention as well, and thus the cost is
\begin{align}
    \mathcal{O}((N^\prime K)^2 d + (N^\prime K)d^2 + N^\prime K N_\mathtt{offset} S d).
\end{align}

In conclusion, the total computational cost of our RAPTR is
\begin{equation}
    \mathcal{O}((d^2 + N_\mathtt{offset}d) \sum_{s=1}^S N_s + (N^2 + (N^\prime K)^2) d + (N + N^\prime K) d^2 + (N +N^\prime K)N_\mathtt{offset} S d).
\end{equation}

\section{Details of Pseudo-3D Deformable Attention}
\label{app:multi_head_attention}
\paragraph{Multi-scale Multi-head Extension of Pseudo-3D Deformable Attention:}
We can extend the pseudo-3D deformable attention defined by Eq.~\ref{eq:pseudo_3d_deformable_attention} in Section~\ref{sec:pseudo_3D_deformable_attention} to multi-scale and multi-head operation.
First, given $M$ heads and $S$ feature scale levels, Eq.~\ref{eq:feature_sampling} is extended as
\begin{equation}
    \fbf_{ms,\mathtt{hor}}^{(i)} = \Fbf_{s,\mathtt{hor}}(x+\Delta x_{msi}, z + \Delta z_{msi}),  \quad
    \fbf_{ms,\mathtt{ver}}^{(i)} = \Fbf_{s, \mathtt{ver}}(y+\Delta y_{msi}, z + \Delta z_{msi}),
\end{equation}
where $\Fbf_{s, \mathtt{hor}}, \Fbf_{s, \mathtt{ver}}$ are the $s$-th level feature maps for horizontal and vertical view, respectively, and $\Delta\{x,y,z\}_{msi}$ is the $i$-th sampling offset in the $m$-th head on the $s$-th level feature.
Subsequently, we collect the sampled features as $\Fbf_{ms,\mathtt{attn}} = \{\fbf_{ms, \mathtt{hor}}^{(1)}, \fbf_{ms, \mathtt{ver}}^{(1)}, \cdots, \fbf_{ms, \mathtt{hor}}^{(N_\mathtt{offset})}, \fbf_{ms, \mathtt{ver}}^{(N_\mathtt{offset})}\}$, and then extend Eq.~\ref{eq:pseudo_3d_deformable_attention} as
\begin{equation}
    \bar{\Fbf}_{\mathtt{attn}} = \sum_{m=1}^{M} \Wbf_{m} [\sum_{s=1}^{S} \sum^{N_{\mathtt{offset}}}_{i=1} (A_{msi,0}  \Wbf_m^\prime \Fbf_{ms, \mathtt{attn}}^{(2i-1)} + A_{msi,1}  \Wbf_m^\prime \Fbf_{ms, \mathtt{attn}}^{(2i)})],
\end{equation}
where $\Wbf_m \in \mathbb{R}^{d\times d_v}$ and $\Wbf_m^\prime \in \mathbb{R}^{d_v\times d}$ with $d_v = d/M$ are learnable weight matrices, and $A_{msi,0}$ and $A_{msi,1}$ are the attention weights of the $i$-th sampled deformable radar feature on the $s$-th level feature in the $m$-th attention head, for the horizontal and vertical radar views.
Attention weights are normalized per head by $\sum_{s}^{S}\sum_{i}^{N_\mathtt{offset}} A_{msi} = 1$.

\paragraph{Implementation of Sampling Location Determination:}
We describe the pseudo-3D deformable attention in Section~\ref{sec:pseudo_3D_deformable_attention} as computing sampling locations by adding $N_\mathtt{offset}$ offsets, which are derived from each query, to a corresponding reference point.
From an implementation point of view, our pose decoder adopts a structurally extended design.
In our implementation, each query corresponds not to a single keypoint, but to the entire pose of a person, represented as $K \times 3$ coordinates.
Accordingly, we replace the notion of a reference point with a reference \textit{pose}, formatted as an $N \times 3K$ tensor for $N$ queries.
Sampling locations are determined by calculating the $3K$ offsets per query and adding them element-wise to the corresponding reference pose $K \times 3$.
In this way, we virtually have $K$ sampling points for each pose.
This effectively enables a single query to take care of $K$ spatial locations, allowing feature aggregation within the context of a unified pose.

On the other hand, the implementation in the joint decoder more aligns with the description in Section~\ref{sec:pseudo_3D_deformable_attention}: a joint query in the joint decoder corresponds to a single joint so that the sampling locations are determined by calculating the $N_\mathtt{offset}$ sampling offsets per query and adding them to the corresponding reference joint.
We set $N_\mathtt{offset}=4$ as listed in Table~\ref{tab:hyper_parameters}.

\paragraph{Computational Complexity Comparison with Decoupled 2D Deformable Attention:}
\begin{figure}
    \centering
    \includegraphics[width=0.7\linewidth]{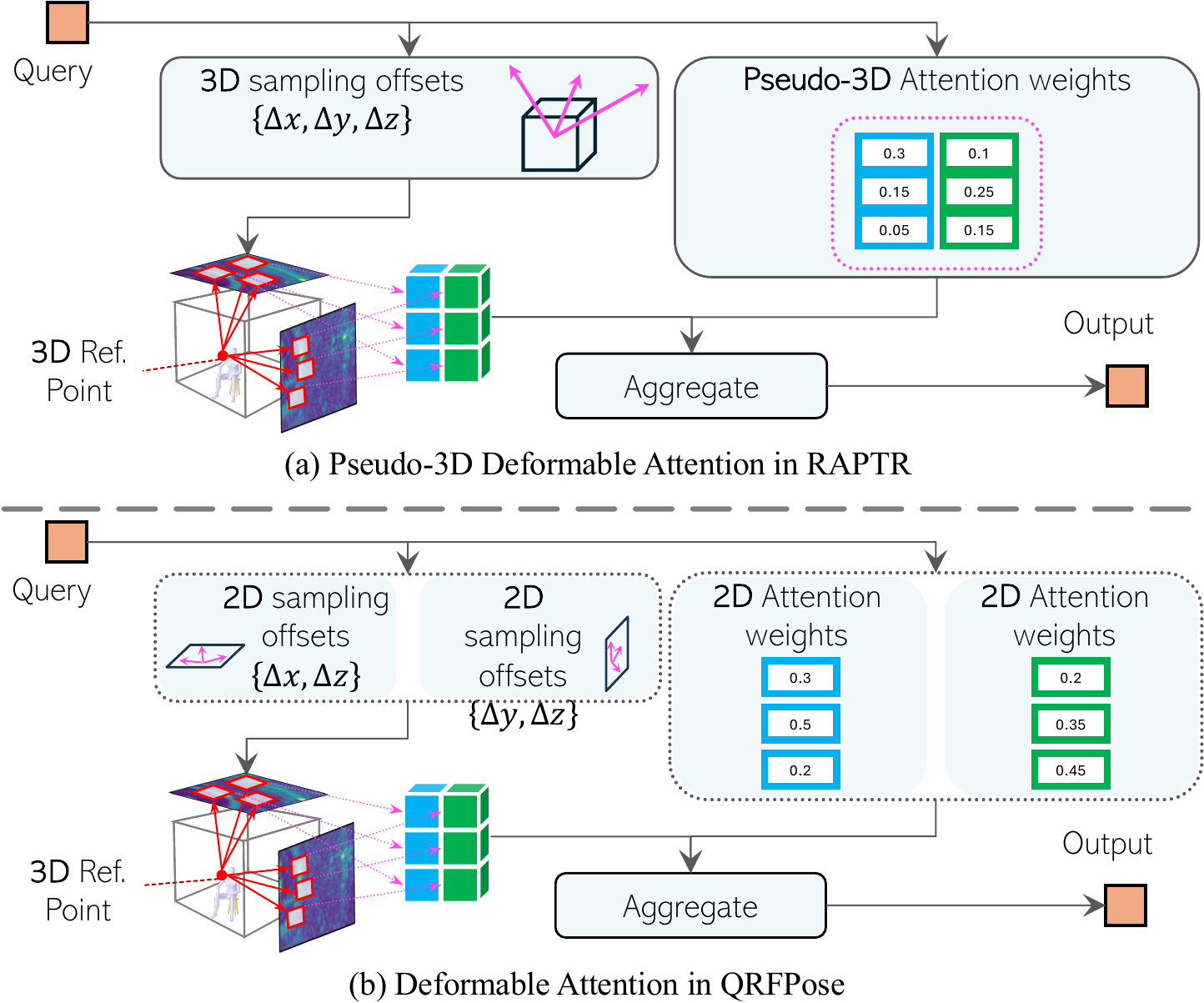}
    \caption{Comparison between (a) pseudo-3D and (b) decoupled 2D deformable attention mechanisms. }
    \label{fig:deform_attn_detail}
\end{figure}
As illustrated in Fig.~\ref{fig:deform_attn_detail}~(a), the pseudo-3D deformable attention adapts a 3D reference point with 3D sampling offsets and the pseudo-3D attention weights are computed over multiple radar views. In comparison, the QRFPose of Fig.~\ref{fig:deform_attn_detail}~(b) adapts a 3D reference point with projected 2D sampling offsets and 2D attention weights are separately computed over each radar view \cite{wan2024qrfpose}. This simple lifting operation may lead to better computational complexity of the pseduo-3D attention over the number of radar views. In the following, we provide a computational complexity analysis for the two types of deformable attention mechanisms, given $V$ radar views:
\begin{itemize}
    \item Decoupled 2D deformable attention: $\mathcal{O}(8VNN_\mathtt{offset}d)$, where
    \begin{itemize}
        \item 3D reference point projected to $V$ 2D radar views: $\mathcal{O}(6VN)$,
        \item Offset estimation: $\mathcal{O}(2VNN_\mathtt{offset}d)$, where $2$ is due to the computation of 2D $(x, y)$ offsets,
        \item Attention weights: $\mathcal{O}(VNN_\mathtt{offset}d)$,
        \item Feature aggregation: $\mathcal{O}(5VNN_\mathtt{offset}d)$, where $5$ is due to bilinear interpolation and weighted sum,
    \end{itemize}
    \item Pseudo-3D deformable attention: $\mathcal{O}(6VNN_\mathtt{offset}d + 3NN_\mathtt{offset}d)$, where
    \begin{itemize}
        \item Offset estimation: $\mathcal{O}(3NN_\mathtt{offset}d)$, where $3$ is due to the computation of the 3D $(x, y, z)$ offsets,
        \item 3D offset projected to $V$ 2D radar views: $\mathcal{O}(6VN)$,
        \item Attention weights: $\mathcal{O}(VNN_\mathtt{offset}d)$,
        \item Feature aggregation: $\mathcal{O}(5VNN_\mathtt{offset}d)$.
    \end{itemize}
\end{itemize}
Note that, in the above analysis, $\mathcal{O}(6VN)$ is excluded from the final complexity expressions as $6VN \ll 5VNN_\mathtt{offset}d$ in practice.
\begin{table}[t]
  \centering
  \caption{Complexity comparison of pseudo-3D vs. decoupled 2D deformable attention.}
  \begin{tabular}{cccccc}
    \toprule
    Queries ($N$) & Views ($V$) & 2D Att & Pseudo-3D Att & Ratio (3D/2D) & Savings \\
    \midrule
    10 & 2  & $160NC$ & $150NC$ & \textbf{0.94}$\downarrow$ & \textbf{6.25\%} \\
    10 & 5  & $400NC$ & $330NC$ & \textbf{0.83}$\downarrow$ & \textbf{17.5\%} \\
    10 & 10 & $800NC$ & $630NC$ & \textbf{0.79}$\downarrow$ & \textbf{21.3\%} \\
    \bottomrule
  \end{tabular}
  \label{tab:attn-cost}
\end{table}
Table~\ref{tab:attn-cost} shows the complexity comparison with a specific number of queries and an increasing number of views. It is observed that the pseudo-3D attention achieves computational savings of $17.5\%$ with 
$V=5$ radar views and $21.3\%$ with 
$V=10$ radar views, compared to the decoupled 2D attention.

\section{Optional View Mask Module}
\label{app:avs}
As an extension of our pseudo-3D deformable attention, we introduce an optional view mask module that aims to put more attention weights on the feature on the more important view with a hard-thresholding approach. 
The view mask module first computes a view selection mask $\Mbf_\mathtt{attn}$ from a query $\qbf$ corresponding to reference points of interest, as:
\begin{align}\label{eq:adaptive_view_selection}
    \Mbf_\mathtt{attn} = \sigma(\lambda \cdot \mathtt{FFN}(\qbf)) \in \mathbb{R}^{N_\mathtt{offset} \times 2},
\end{align}
where $\sigma$ is the Sigmoid function, and $\lambda (\approx 1\mathrm{e}5) \in \mathbb{R}$ makes the sigmoid output very close to 0 or 1 while preserving gradients.
The element $m_{i,j} (\approx 1)$ signals that the $j$th view ($j=0$ to the horizontal and $j=1$ to the vertical) should retain its share of attention at the $i$th sampling point, whereas $m_{i,j} (\approx 0)$ marks it for suppression.

For each $i$th sampling point, we can consider three patterns to adjust the attention weights $f_\mathtt{attn}$ for that point $A_{i,0}, A_{i,1}$ and, potentially, other sampling points according to the corresponding row in the view selection mask $[m_{i,0}, m_{i,1}]$.
\begin{itemize}
    \item $[m_{i,0}, m_{i,1}] = [1,1]$: use both views, weights unchanged;
    \item $[m_{i,0}, m_{i,1}] = [0,1]$ or $[m_{i,0}, m_{i,1}] = [1,0]$: ignore one view and transfer its weight to the other, e.g., when $[m_{i,0}, m_{i,1}] = [0,1]$, the adjusted attention weights are $\hat{A}_{i,0} = A_{i,0} + A_{i,1}, \hat{A}_{i,1} = 0$, to to ensure that the view selection decision at one sampling point does not influence the other sampling points;
    \item $[m_{i,0}, m_{i,1}] = [0,0]$: ignore this sampling point in both views. Its weight is evenly redistributed so that $\sum_{i,j}\hat{A}_{i,j} = 1$ still holds.
\end{itemize}

In this way, the view selection mask adaptively changes through training so that the attention mechanism associates with the view that is more informative for each sampling point.
We can also utilize this view selection module to control the use of multi-view features by manually setting the values in the view selection mask, which we conduct a relating ablation study in the following Appendix~\ref{app:ablation}.

\section{Details of Loss Functions}
\label{app:loss}
\begin{figure}
    \begin{minipage}[b]{0.45\hsize}
        \centering
        \begin{minipage}[b]{0.45\hsize}
            \includegraphics[width=\linewidth]{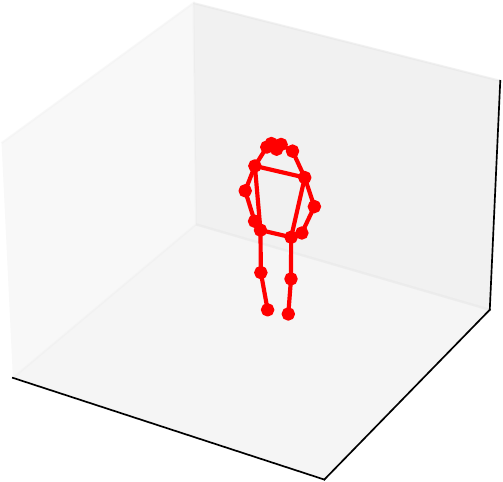}
        \end{minipage}
        \begin{minipage}[b]{0.45\hsize}
            \includegraphics[width=\linewidth]{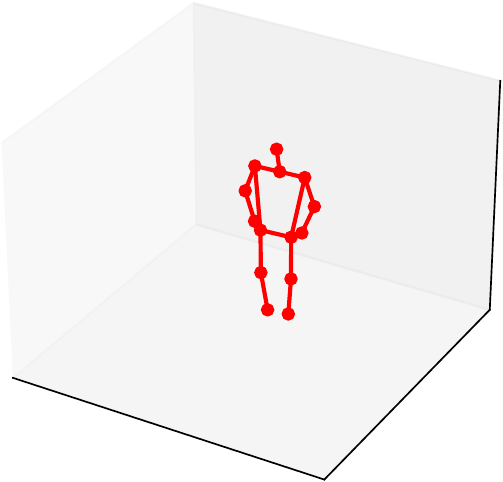}
        \end{minipage}
        \caption{Template keypoints $\Kbf_\mathtt{world}$ for MMVR (left) and HIBER (right) dataset.}
        \label{fig:template_pose}
    \end{minipage}
    \hspace{0.04\hsize}
    \begin{minipage}[b]{0.45\hsize}
        \centering
        \includegraphics[width=0.9\linewidth]{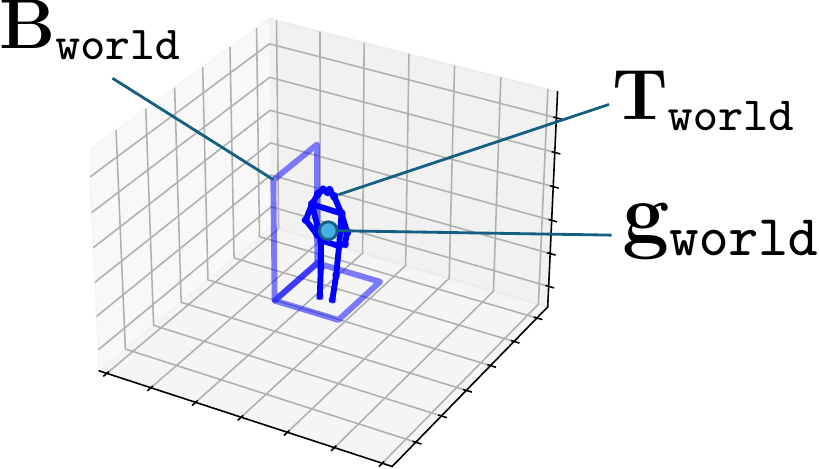}
        \caption{An example of a template pose located in the 3D world coordinate.}
        \label{fig:template_pose_locate}    
    \end{minipage}
\end{figure}

In this section, we give a supplementary explanation of the loss functions in RAPTR provided in Section~\ref{sec:methodology}.
For simplicity, we consider a single corresponding sample from each set, denoted as $\{\bbf_{\mathtt{world}} \in \Bbf_{\mathtt{world}},  \pbf_{\mathtt{image}} \in \Pbf_{\mathtt{image}}, \hat{\pbf}_{\mathtt{image}} \in \hat{\Pbf}_{\mathtt{image}}, \tilde{\pbf}_{\mathtt{world}} \in \tilde{\Pbf}_{\mathtt{world}}, \hat{\pbf}_{\mathtt{world}} \in \hat{\Pbf}_{\mathtt{world}}\}$.

\paragraph{Coarse-grained 3D Loss:}
For 3D gravity loss, we compute the centroid of a 3D BBox label in $\bbf_\mathtt{world} = [x_\mathtt{min}, y_\mathtt{min}, z_\mathtt{min}, x_\mathtt{max}, y_\mathtt{max}, z_\mathtt{max}]$ as the 3D gravity center label $\gbf_\mathtt{world} \in \Rset^{1 \times 3}$ as
\begin{align}
    \gbf_\mathtt{world} = [\frac{x_\mathtt{max} - x_\mathtt{min}}{2}, \frac{y_\mathtt{max} - y_\mathtt{min}}{2}, \frac{z_\mathtt{max} - z_\mathtt{min}}{2}].
\end{align}
Given a refined 3D pose estimate $\hat{\pbf}_\mathtt{world} = \{(\hat{\pbf}_{\mathtt{world}, x}^{(k)}, \hat{\pbf}_{\mathtt{world}, y}^{(k)}, \hat{\pbf}_{\mathtt{world}, z}^{(k)})\}_{k=1}^{K}$ at the joint decoder, we also collapse it into its centroids as $\hat{\gbf}_\mathtt{world} \in \Rset^{1 \times 3}$ as
\begin{align}
    \hat{\gbf}_\mathtt{world} = [\frac{1}{K}\sum_k \hat{\pbf}_{\mathtt{world}, x}^{(k)}, \frac{1}{K}\sum_k \hat{\pbf}_{\mathtt{world}, y}^{(k)}, \frac{1}{K}\sum_k \hat{\pbf}_{\mathtt{world}, z}^{(k)}],
\end{align}
where $\hat{\pbf}_{\mathtt{world}, x/y/z}^{(k)}$ is the $x$-, $y$-, and $z$-coordinate for the $k$-the joint.
The 3D gravity loss $\mathcal{L}_\mathtt{gravity}$ is defined as the Euclidean distance between the two 3D gravity centers, $\gbf_\mathtt{world}$ and $\hat{\gbf}_\mathtt{world}$.

For 3D template loss, we construct a 3D template keypoint label for each $\bbf_\mathtt{world}$ using template keypoints at the coordinate origin $\Kbf_\mathtt{world} \in \Rset^{K \times 3}$.
The template pose $\Tbf_\mathtt{world} \in \Rset^{K \times 3}$ is given as $\Tbf_\mathtt{world} = \Kbf_\mathtt{world} + \mathbf{1}^{\top} \gbf_\mathtt{world}$.
Fig.~\ref{fig:template_pose} shows the template keypoints $\Kbf_\mathtt{world}$ with different numbers of keypoints for the MMVR and HIBER datasets, and Fig.~\ref{fig:template_pose_locate} provides an example of locating these template keypoints in the 3D world coordinate based on the location of a 3D BBox label.
The 3D template loss $\mathcal{L}_\mathtt{template}$ is defined as the Euclidean distance between the template poses $\Tbf_\mathtt{world}$ and the initial 3D pose estimates $\tilde{\Pbf}_\mathtt{world}$ from the pose decoder.

\paragraph{Fine-grained 2D Loss:}
Specifically in the fine-grained 2D loss, OKS loss $\mathcal{L}_\mathtt{OKS}$ is the loss function based on object keypoint similarity~(OKS), a metric used to evaluate the accuracy of keypoint estimations taking into account the object scale and keypoint visibility, which is defined as
\begin{align}
    \mathrm{OKS}(\pbf_\mathtt{image}, \hat{\pbf}_\mathtt{image}) = \sum_k \exp(-\frac{d_k^2}{2s^2 \psi_k^2}),
\end{align}
where $d_k$ is the distance between the $k$-th estimated joint $\hat{\pbf}_\mathtt{image}^{(k)}$ and the corresponding label $\pbf_\mathtt{image}^{(k)}$, $s$ is the object scale, and $\psi_k$ is a pre-defined constant for the $k$-th joint.
Here, we assume that all keypoint labels are annotated as visible points.
Since OKS is a metric in which higher values indicate greater similarity, its negative logarithm $-\log(\mathrm{OKS})$ is taken when used as a loss function $\mathcal{L}_\mathtt{OKS}$.

\section{Datasets}
\label{app:dataset}
\paragraph{HIBER:}
HIBER~\cite{RFMask23} is an open-source multi-view radar dataset for indoor human perception tasks including detection, segmentation, and keypoint estimation.
They provide horizontal and vertical radar heatmaps and corresponding labels such as 2D BBoxes, 2D segmentation masks, 2D keypoints, 3D BBoxes, and 3D keypoints.
Among its data environments, WALK and MULTI are currently available.
WALK comprises frames that feature a single individual, while frames in MULTI consistently depict two individuals walking concurrently.
The frames provided are captured from ten distinct viewpoints within a single room, designated as ``view01'' through ``view10.''
We use ``view02'' to ``view10'' for training, validating, and testing the models, with the data splits provided, and the specific number of frames is listed in Table~\ref{tab:hyper_parameters}.

\paragraph{MMVR:}
MMVR~\cite{Rahman2024_mmvr} is a more recent open-source multi-view radar dataset for indoor human perception.
They provide horizontal and vertical radar heatmaps and corresponding labels, such as 2D BBoxes, 2D segmentation masks, 2D keypoints, and 3D BBoxes.
They collected data from 25 subjects in 6 different scenarios (e.g., open/cluttered office spaces) spanning over 9 days.
MMVR consists of 1) P1: single-person scenarios in an open space without any obstacles, and 2) P2: multi-person scenarios in a cluttered office spaces, including sitting postures.
P1 is designed to establish fundamental benchmarks for radar-based human perception tasks, while P2 is designed to challenge with more realistic and complicated indoor scenarios and cross-environment, cross-subject generalization.
The data split we use is S1 that they provide, and the specific number of frames is listed in Table~\ref{tab:hyper_parameters}.

\section{Hyper Parameters}
\label{app:training_parameter}
In the evaluations presented in Section~\ref{sec:evaluation}, the model training for all baselines and our RAPTR and its variants share the hyper parameters outlined in Table~\ref{tab:appendix_training_parameters}, unless otherwise specified.

\begin{table}[t]
    \scriptsize
    \centering
    \caption{Hyper parameters for RAPTR}
    \label{tab:hyper_parameters}
    \begin{tabular}{lccc} \toprule
         \multirow{2}{*}{\textbf{Name}} & \multirow{2}{*}{\textbf{Notation}} & \multicolumn{2}{c}{\textbf{Value}} \\
         \cline{3-4}
         & & \rule{0pt}{9pt}\textbf{HIBER} & \textbf{MMVR} \\ 
         \midrule
         \textbf{Data} & & \\
         Radar image resolution & $W, H, D$ & 160, 160, 200 & 128, 128, 256 \\
         \# of training samples & - & 59000 / 54280 (WALK / MULTI) & 86579 / 190441(P1S1 / P2S1)\\
         \# of validation samples & - & 6490 / 5900 (WALK / MULTI) & 10538 / 23899 (P1S1 / P2S1)\\
         \# of test samples & - & 3540 / 3540 (WALK / MULTI) & 10785 / 23458 (P1S1 / P2S1) \\
         \# of keypoints & $K$ & 14 & 17 \\
         \midrule
         \textbf{Model params} & & \\
         Backbone & - & \multicolumn{2}{c}{ResNet 18} \\
         \# of feature scale & $S$ & \multicolumn{2}{c}{3} \\
         Feedforward dimension in Transformer & - & \multicolumn{2}{c}{1024} \\
         \# of encoder layers & $L_\mathtt{enc}$ & \multicolumn{2}{c}{3} \\
         \# of pose decoder layers & $L_\mathtt{pose}$ & \multicolumn{2}{c}{2} \\
         \# of joint decoder layers & $L_\mathtt{joint}$ & \multicolumn{2}{c}{3} \\
         \# of deformable sampling offsets & $N_\mathtt{offset}$ & \multicolumn{2}{c}{$K$ / 4 (pose / joint decoder)}\\
         \# of heads in multi-head attention & - & \multicolumn{2}{c}{8} \\
         Feature dimension & $d$ & \multicolumn{2}{c}{128} \\
         \# of input frames & $T$ & \multicolumn{2}{c}{4} \\
         \# of pose query & $N$ & \multicolumn{2}{c}{10} \\
         \midrule
         \textbf{Training params} & & \\
         Optimizer & - & \multicolumn{2}{c}{AdamW} \\
         Base learning rate & - & \multicolumn{2}{c}{2e-4} \\
         Weight decay & - & \multicolumn{2}{c}{1e-4} \\
         LR scheduler & - & \multicolumn{2}{c}{Cosine Decay} \\
         Batch size & - & \multicolumn{2}{c}{32} \\
         Epochs & - & \multicolumn{2}{c}{50} \\
         Gradient clip norm & - & \multicolumn{2}{c}{0.1} \\
         Early stopping patience & - & \multicolumn{2}{c}{5 epochs} \\
         \midrule
         \textbf{Loss weights} & & \\
         3D template loss & $\lambda_1$ & \multicolumn{2}{c}{1.0} \\
         3D gravity loss & $\lambda_2$ & \multicolumn{2}{c}{1.0} \\
         2D Keypoint loss & $\lambda_3$ & \multicolumn{2}{c}{5.0} \\
         2D OKS loss & $\lambda_4$ & \multicolumn{2}{c}{1.0} \\
         Class loss & $\lambda_5$ & \multicolumn{2}{c}{1.0} \\
         \midrule
         \textbf{Computational Resource} & & \\
         GPU & & \multicolumn{2}{c}{NVIDIA A40} \\
         \# of workers & & \multicolumn{2}{c}{8} \\
         Approximate training time & & \multicolumn{2}{c}{3 hours / 10K samples}\\
         \bottomrule
         
    \end{tabular}
    \label{tab:appendix_training_parameters}
\end{table}

\section{Baseline Implementations}
\label{app:baseline}

We provide the specific implementation for the baseline methods that we use in the evaluation.

\paragraph{Person-in-WiFi 3D:}
We refer to the official implementation~\cite{yan2024person} and modify some parts of the code to make them compatible with the datasets that we use.
We employ a ResNet backbone to extract multi-scale features from the radar heatmaps, as well as our RAPTR does.
We then take the C4 feature map, flatten it as the $N_\mathtt{token}$ tokens with dimensions of $d$, and feed it into the network.
Specifically, $N_\mathtt{token} = 260$ for the HIBER dataset and $256$ for the MMVR dataset. 
Since the original study uses Wi-Fi channel state information~(CSI), which inherently lacks explicit spatial structure, and transforms it into 180 tokens for input, our approach of converting C4 feature maps derived from radar heatmaps into approximately 200 tokens can be reasonably justified in terms of fairness and comparability.
Regarding the loss function, we implement the loss as the summation of class loss, 2D keypoint loss, refined 2D keypoint loss, and 3D gravity loss, with loss weights of 1.0, 5.0, 10.0, and 1.0, respectively.

\paragraph{HRRadarPose:}
We refer to the official implementation~\cite{yuan2024rtpose} and modify some parts of the code to make them compatible with the datasets that we use.
First, we exclusively utilize horizontal view radar heatmaps, excluding vertical view heatmaps.
This decision stems from the disparity in angular resolution between the elevation and azimuth axes reported in the HRRadarPose paper.
The resolution of the elevation axis is 18 degrees, while the resolution of the azimuth axis is 1.4 degrees, and only the azimuth resolution is comparable to that of HIBER and MMVR (1.3 degrees).
We presume that we could solely use the horizontal view heatmaps while ensuring fairness in our evaluations.
In addition, we expand the original codes to multi-person scenarios by implementing Non-Maximum Suppression~(NMS) on the predictions.
Regarding the loss function, we implement the loss as the summation of heatmap loss, 2D keypoint loss, and 3D gravity loss, with loss weights of 5.0, 1.0, and 1.0, respectively.

\paragraph{QRFPose:}
Currently, the authors of the paper have not released official codes.
Therefore, we independently replicated the implementation based on the architectures and parameters outlined in the paper.
We verify that our implementation replicates performance similar to that of the original report using 3D keypoint labels.
To ensure a fair comparison, we set the number of Transformer decoder layers to 5, which is equivalent to the total number of layers in the pose decoder and the joint decoder in our RAPTR model.
Although the original implementation uses RLE loss~\cite{li2021human} as the keypoint regression loss function, we employ the conventional Euclidean distance loss in our implementation so that we can integrate the loss with the 3D gravity loss in a more balanced way.
Regarding the loss function, we implement the loss as the summation of class loss, 2D keypoint loss, and 3D gravity loss, with loss weights of 1.0, 5.0, and 1.0, respectively.

\section{Metrics}
\label{app:metrics}
For simplicity, we omit the subscripts that indicate the coordinate system in which the keypoints or BBoxes are defined ($\mathtt{radar}, \mathtt{world}$) in this section.
In addition, $\pbf \in \Pbf$, $\bbf \in \Bbf$, and $\hat{\pbf} \in \hat{\Pbf}$ denote the corresponding samples taken from the 3D keypoint labels, the 3D BBox labels, and the 3D pose estimates, respectively.

\paragraph{MPJPE:}
We employ Mean Per Joint Position Error~(MPJPE) as the performance metric to evaluate the 3D pose estimation capabilities of the models.
Given a 3D keypoint label $\pbf = \{\pbf^{(k)} | (x^{(k)}, y^{(k)}, z^{(k)})\}_{k=1}^{K}$ and the corresponding estimate $\hat{\pbf} = \{\hat{\pbf}^{(k)} | (\hat{x}^{(k)}, \hat{y}^{(k)}, \hat{z}^{(k)})\}_{k=1}^{K}$.
MPJPE is defined as:
\begin{align}
    \mathrm{MPJPE} = \frac{1}{K} \sum_{k=1}^{K} \|\pbf^{(k)} - \hat{\pbf}^{(k)}\|_2.
\end{align}
The unit for MPJPE that we use is the centimeter in the world coordinate system.
We also evaluate MPJPE along each axis: horizontal (h), vertical (v), and depth (d).

\paragraph{3D BBox-based Metrics for MMVR:}
For the evaluation of the MMVR dataset, due to the absence of 3D keypoint labels in the dataset, we approximate the pose estimation performance in a different way from that for the HIBER dataset.
Specifically, we calculate 1) the distance between the center of the 3D pose estimate $\hat{\Pbf}$ and the 3D \ac{BBox} label $\Bbf$, and 2) the absolute error in the edge lengths along each axis of the box.
Specifically, we first construct a 3D \ac{BBox} that encloses the estimated 3D keypoints as $\hat{\bbf} = [\min(\pbf_{x}), \min(\pbf_{y}), \min(\pbf_{z}), \max(\pbf_{x}), \max(\pbf_{y}), \max(\pbf_{z})]$ where $\pbf_{x}, \pbf_{y}, \pbf_{z}$ is the set of $x$-, $y$- and $z$-coordinates of the estimated keypoints.
We then calculate the center coordinate of the 3D BBox label $\bbf = [x_\mathtt{min}, y_\mathtt{min}, z_\mathtt{min}, x_\mathtt{max}, y_\mathtt{max}, z_\mathtt{max}]$ and $\hat{\bbf}$ as
\begin{align}
    \gbf &= [\frac{x_\mathtt{max} - x_\mathtt{min}}{2}, \frac{y_\mathtt{max} - y_\mathtt{min}}{2}, \frac{z_\mathtt{max} - z_\mathtt{min}}{2}], \nonumber \\
    \hat{\gbf} &= [\frac{\max(\pbf_{x}) - \min(\pbf_{x})}{2}, \frac{\max(\pbf_{y}) - \min(\pbf_{y})}{2}, \frac{\max(\pbf_{z}) - \min(\pbf_{z})}{2}].
\end{align}
The center distance between the BBoxes is the Euclidean distance between $\gbf$ and $\hat{\gbf}$.
We also calculate the edge lengths of $\bbf$ and $\hat{\bbf}$ as
\begin{align}
    \lbf &= (x_\mathtt{max} - x_\mathtt{min}, y_\mathtt{max} - y_\mathtt{min}, z_\mathtt{max} - z_\mathtt{min}), \nonumber \\
    \hat{\lbf} &= (\max(\pbf_{x}) - \min(\pbf_{x}), \max(\pbf_{y}) - \min(\pbf_{y}), \max(\pbf_{z}) - \min(\pbf_{z})),
\end{align}
and we calculate the absolute error of the edge length along each axis.

\section{Additional Ablation Studies and Visualization}\label{app:ablation}
To validate the effectiveness of RAPTR, we conduct additional ablation studies.
Unless otherwise specified, we conduct the studies with the hyper parameters in Table~\ref{tab:hyper_parameters}.

\subsection{Numerical Results}
\paragraph{Additional Results for MMVR on P2S1:}
Table~\ref{tab:app_result_mmvr} shows the evaluation results for the MMVR dataset on P2S1.
P2S1 includes cluttered indoor scenarios with multiple subjects, which is thus more challenging than P1S1.
RAPTR outperforms baselines and shows improvements in center distance by 71.54\%, 85.28\%, and 69.47\% compared to Person-in-WiFi 3D, QRFPose, and HRRadarPose, respectively.
We defer the qualitative evaluation for MMVR P2S1 to Appendix~\ref{app:qualitative_results}.
\begin{table}[ht]
    \centering
    \scriptsize
    \setlength\tabcolsep{3.3pt}
    \caption{3D pose estimation performance on MMVR (P2S1).}
    \label{tab:app_result_mmvr}
    \begin{tabular}{c|cccc} \toprule
         \multirow{2}{*}{\textbf{Method}} & \multirow{2}{*}{\textbf{Center distance (\si{\centi\meter})}} & \multicolumn{3}{c}{\textbf{Edge length error (\si{\centi\meter})}} \\
         \cline{3-5}
         & & (h) & (v) & (d) \\ \midrule
        Person-in-WiFi 3D & 103.43 & 48.29 & 112.75 & 152.88 \\
        QRFPose & 200.03 & 115.80 & 126.14 & 335.30 \\
        HRRadarPose & 96.43 & 32.19 & 51.02 & 175.04 \\ \midrule
        RAPTR (ours) & \cellcolor{gray!20}\textbf{29.44} & \cellcolor{gray!20}\textbf{18.74} & \cellcolor{gray!20}\textbf{27.14} & \cellcolor{gray!20}\textbf{40.29} \\ \bottomrule
    \end{tabular}
\end{table}

\paragraph{Full 3D Supervision:}
We compare the performance of our RAPTR under (i) full supervision with fine-grained 3D keypoint labels and (ii) weak supervision with 2D keypoint and 3D BBox labels for HIBER (MULTI).
Table~\ref{tab:app_result_hiber_3d} shows the performance comparison in MPJPE.
Under full 3D supervision, the RAPTR architecture achieves an MPJPE of \SI{8.93}{\centi\meter}.
Even when trained under weak supervision, MPJPE increases by only about \SI{10}{\centi\meter}, indicating that our structured loss design with two-stage decoding approach effectively learns reliable 3D body structures.
\begin{table}[htb]
    \scriptsize
    \centering
    \setlength\tabcolsep{3pt}
    \caption{RAPTR performance with full and weak supervision (HIBER MULTI).}
    \label{tab:app_result_hiber_3d}
    \begin{tabular}{c|cccccccc|cccc} \toprule
        \textbf{Method} & \textbf{Head} & \textbf{Neck} & \textbf{Shoulder} & \textbf{Elbow} & \textbf{Wrist} & \textbf{Hip} & \textbf{Knee} & \textbf{Ankle} & \textbf{MPJPE} & \textbf{(h)} & \textbf{(v)} & \textbf{(d)} \\ \midrule
        RAPTR (Full 3D supervision) & 7.90 & 5.28 & 6.41 & 7.94 & 12.43 & 5.75 & 8.80 & 14.65 & 8.93 & 4.97 & 2.86 & 5.12 \\
        RAPTR (Weak supervision) & 18.39 & 13.13 & 16.44 & 20.12 & 24.62 & 15.01 & 17.76 & 23.22 & 18.99 & 7.80 & 4.38 & 14.54 \\
         \bottomrule
    \end{tabular}
\end{table}

\paragraph{Effect of 3D Templates and Their Scales:}~We evaluate the impact of 3D templates and their scale on the final MPJPE performance. Specifically, we experiment with
\begin{itemize}
    \item A \textbf{standing pose} scaled by two factors: $0.5\times$ and $1\times$,
    \item A \textbf{sitting pose} of a \SI{1.6}{\meter}-tall person,
    \item A \textbf{learned scaling factor} applied to the standard standing pose.
\end{itemize}
\begin{table}[t]
    \centering
    \small
    \captionsetup{labelfont={color=black},font={color=black}}
    \caption{Effect of 3D Templates and their scales on the performance (HIBER MULTI).}
    \begin{tabular}{ll} \toprule
         \textbf{3D Template} & \textbf{MPJPE} \\
         \midrule
         Standing (scale=1) & 18.99 $\pm$ 0.16 \\
         Standing (scale=0.5) & 20.11 $\pm$ 0.54 \\
         Sitting (scale=1) & 20.84 $\pm$ 0.91 \\
         Standing (learned scale) & 23.13 $\pm$ 0.33 \\
         \bottomrule
    \end{tabular}
    \label{tab:app_template_impact}
\end{table}
Table~\ref{tab:app_template_impact} suggests that the choice of 3D template has minor impacts on the final MPJPE, likely due to the refinement capability of the second-stage joint decoder, as long as the first-stage decoder generates a reasonable, human-like initial pose.

\paragraph{Effect of Loss Weighting Factors:}
We assess the RAPTR performance under varying loss weighting factors $\lambda_i$. Three configurations are evaluated: 1) equal weights for all loss terms, 2) increased weights on 3D losse terms,  and 3) increased weights on 2D keypoint loss.
Table~\ref{tab:app_loss_weights} provides the performance comparison among these settings.

When all weighting factors are set to $1.0$, RAPTR achieves an average MPJPE o \SI{19.91}{\centi\meter}. Increasing the weights of the 3D losses to $5.0$ degrades performance, resulting in an MPJPE of \SI{24.04}{\centi\meter}.
In contrast, emphasizing the 2D keypoint loss yields the best performance with an MPJPE of \SI{18.99}{\centi\meter}. 
These results suggest that appropriately balancing the loss terms, particularly by increasing the weight of the 2D keypoint loss, plays a crucial role in enhancing joint localization accuracy.

\begin{table}[htb]
    \centering
    \scriptsize
    \captionsetup{labelfont={color=black},font={color=black}}
    \caption{Effect of loss weighting factors on the RAPTR performance (HIBER MULTI).}
    \label{tab:app_loss_weights}
    \begin{tabular}{ccccc|c} \toprule
        $\lambda_1$ & $\lambda_2$ & $\lambda_3$ & $\lambda_4$ & $\lambda_5$ & \multirow{2}{*}{\textbf{MPJPE}} \\
        3D template & 3D gravity & 2D keypoint & 2D OKS & class & \\
        \midrule
        1.0 & 1.0 & 1.0 & 1.0 & 1.0 & 19.91 $\pm$ 0.65 \\
        5.0 & 5.0 & 1.0 & 1.0 & 1.0 & 24.04 $\pm$ 0.75 \\
        1.0 & 1.0 & 5.0 & 1.0 & 1.0 & \textbf{18.99 $\pm$ 0.16} \\
        \bottomrule
    \end{tabular}
\end{table}
\paragraph{Impact of View Selection:}
To investigate how the use of features from horizontal and vertical views affects performance, we evaluate the performance of 3D pose estimation by configuring the view selection mask $\Mbf_\mathtt{attn} \in \mathbb{R}^{N_\mathtt{offset} \times 2}$, integrated into pseudo-3D deformable attention as described in Appendix~\ref{app:avs}, according to several predefined patterns.
Specifically, given that the first column of the view selection mask $\Mbf_\mathtt{attn}[:, 0]$ and the second column $\Mbf_\mathtt{attn}[:, 1]$ correspond to horizontal and vertical views, respectively, we set the values in the mask as
\begin{itemize}
    \item \textbf{Both Views}: all values in $\Mbf_\mathtt{attn}$ to 1 so that the attention weight matrix $f_\mathtt{attn}$ is used as is;
    \item \textbf{Horizontal View Only}: $\Mbf_\mathtt{attn}[:, 0] = \mathbf{1}$ and $\Mbf_\mathtt{attn}[:, 1] = \mathbf{0}$ so that the features sampled from the horizontal feature map are aggregated and those of the vertical feature are omitted;
    \item \textbf{Vertical View Only}: The reversed one of ``horizontal only'': $\Mbf_\mathtt{attn}[:, 0] = \mathbf{0}$ and $\Mbf_\mathtt{attn}[:, 1] = \mathbf{1}$;
    \item \textbf{Random Mask}: The values in the mask are randomly assigned for each mini-batch step in the training process;
    \item \textbf{Adaptive View Selection}: The mask values are adaptively determined by the corresponding queries, described in Appendix~\ref{app:avs}.
\end{itemize}
\begin{table}[htb]
    \centering
    \scriptsize
    \captionsetup{labelfont={color=black},font={color=black}}
    \caption{Effect of view selection patterns on the performance. }
    \label{tab:app_view_selection}
    \begin{tabular}{c|cccc} \toprule
         & \textbf{MPJPE} & \textbf{(h)} & \textbf{(v)} & \textbf{(d)} \\
        \midrule
        Both Views & 20.31 $\pm$ 0.34 & 8.44 $\pm$ 0.34 & 5.02 $\pm$ 0.05 & 15.21 $\pm$ 0.67 \\
        Horizontal View Only & 20.55 $\pm$ 2.05 & 8.74 $\pm$ 0.67 & 5.42 $\pm$ 0.82 & 14.97 $\pm$ 1.99 \\
        Vertical View Only & 23.78 $\pm$ 0.77 & 12.24 $\pm$ 1.45 & 4.66 $\pm$ 0.14 & 16.61 $\pm$ 1.15 \\
        Random Mask & 21.18 $\pm$ 0.33 & 8.75 $\pm$ 0.78 & 4.96 $\pm$ 0.72 & 16.04 $\pm$ 0.10 \\
        \midrule
        Adaptive View Selection (ours) & 18.99 $\pm$ 0.16 & 7.80 $\pm$ 0.31 & 4.38 $\pm$ 0.25 & 14.54 $\pm$ 0.13 \\
        \bottomrule
    \end{tabular}
\end{table}
Table~\ref{tab:app_view_selection} shows the performance comparison in MPJPE.
When using only a single view as input, restricting the input to either the horizontal or vertical view leads to a noticeable increase in MPJPE, averaging \SI{20.55}{\centi\meter} and \SI{23.78}{\centi\meter}, respectively.
In contrast, under the multi-view input setting, the adaptive view selection strategy achieves the lowest average MPJPE of \SI{18.99}{\centi\meter}, outperforming both the “random mask” and “both views” configurations.

\paragraph{Analysis of Attention Weight Assignment:}
We investigate the contribution of each radar view by analyzing the distribution of attention weights and view-selection patterns in the pose and joint decoders.
Table~\ref{tab:app_attention_weight} shows the joint-wise breakdown of attention weight assignment.
Note that weights in the pose decoder are normalized over all joints and views, while those in the joint decoder are normalized per joint. Due to averaging over the test set, values may not sum to exactly 1.
In the pose decoder, both horizontal and vertical views receive relatively balanced attention across joints.
On the other hand, in the joint decoder, vertical views consistently receive higher weights.
Specifically, the edge joints exhibit larger disparities, such as 25.07\% difference for the knee and 18.39\% for the wrist.
Since the pose decoder merely aligns the estimates to template poses, it only requires a rough estimate of the overall pose structure, and thus the attention weights are almost evenly distributed for both views.
In contrast, the joint decoder is tasked with refining each joint, which makes the vertical view more critical, as it provides more comprehensive visibility across all joints.
\begin{table}[htb]
    \centering
    \scriptsize
    \caption{Attention weight assignment in the pose/joint decoders ($\times$1e-2). The larger value indicates the more attention is weighted on the view for each joint. While the pose decoder assigns balanced attention across joints, the joint decoder makes the vertical view more critical with larger weights.}
    \label{tab:app_attention_weight}
    \begin{tabular}{c|ccccccccc} \toprule
        & & \textbf{Head} & \textbf{Neck} & \textbf{Shoulder} & \textbf{Elbow} & \textbf{Wrist} & \textbf{Hip} & \textbf{Knee} & \textbf{Ankle} \\
        \midrule
        \multirow{2}{*}{Pose} & Horizontal & \textbf{4.59} & 3.07 & \textbf{3.53} & 3.07 & 2.82 & 3.56 & 2.30 & 1.57\\
         & Vertical   & 4.01 & \textbf{3.35} & 2.33 & \textbf{3.74} & \textbf{3.68} & \textbf{4.68} & \textbf{2.36} & \textbf{3.63} \\
        \midrule
        \multirow{2}{*}{Joint} & Horizontal & 34.08 & 40.32 & 41.96 & 40.18 & 31.45 & 38.40 & 32.06 & 35.72\\
         & Vertical   & \textbf{50.20} & \textbf{46.83} & \textbf{46.31} & \textbf{46.84} & \textbf{49.84} & \textbf{49.94} & \textbf{57.13} & \textbf{47.69} \\
        \bottomrule
    \end{tabular}
\end{table}

In RAPTR, we adopt a view selection approach in which the view selection mask is adaptively determined by the corresponding queries, as described in Appendix~\ref{app:avs}, and Table~\ref{tab:app_view_selection_dist} shows the joint-wise breakdown of view selection assignments in the pose and joint decoders.
``Omit Both'', ``Use Horizontal'', ``Use Vertical'', and ``Use Both'' are represented by the view selection mask values $[0,0]$, $[1,0]$, $[0,1]$, and $[1,1]$ for each corresponding row in the mask, respectively.
While the pose decoder shows no prominent value imbalance across the assigned patterns, the joint decoder tends to rely more on both views, with high ``Use Both'' ratios observed for almost all joints, such as 31.06\% for shoulder, 32.48\% for hip, and 32.44\% for neck.
This trend reflects that the joint decoder prefers to integrate information from both views when refining keypoint locations.
\begin{table}[htb]
    \centering
    \scriptsize
    \caption{The joint-wise breakdown of view-selection assignments in the pose/joint decoders (\%). The joint decoder relies on both views more than the pose decoder, especially for edge joints.}
    \label{tab:app_view_selection_dist}
    \begin{tabular}{c|ccccccccc} \toprule
        & & \textbf{Head} & \textbf{Neck} & \textbf{Shoulder} & \textbf{Elbow} & \textbf{Wrist} & \textbf{Hip} & \textbf{Knee} & \textbf{Ankle} \\
        \midrule
        \multirow{4}{*}{Pose} & Omit Both & 25.81 & 21.09 & 23.59 & 22.50 & 26.16 & 18.99 & 25.31 & \textbf{28.42} \\
        & Use Horizontal & 22.61& 25.47 & \textbf{27.15} & \textbf{31.84} & 22.74 & 23.29 & 19.35 & 26.55 \\
        & Use Vertical & \textbf{29.65} & 26.70 & 22.72 & 20.79 & 24.43 & 21.76 & 22.42 & 23.33 \\
        & Use Both & 21.92 & \textbf{26.74} & 26.54 & 24.86 & \textbf{26.67} & \textbf{35.96} & \textbf{32.92} & 21.70 \\
        \midrule
        \multirow{4}{*}{Joint} & Omit Both & 20.73 & 17.87 & 19.94 & 19.24 & 24.20 & 21.54 & 19.13 & 23.48 \\
        & Use Horizontal & 22.75 & 21.55 & 22.94 & 23.18 & 21.24 & 25.86 & 23.17 & 25.25 \\
        & Use Vertical & 27.99 & 28.14 & 26.06 & 27.69 & \textbf{28.20} & 20.12 & 26.52 & 24.98 \\
        & Use Both & \textbf{28.53} & \textbf{32.44} & \textbf{31.06} & \textbf{29.88} & 26.37 & \textbf{32.48} & \textbf{31.18} & \textbf{26.29} \\
        \bottomrule
    \end{tabular}
\end{table}
\subsection{Additional Visualizations}\label{app:qualitative_results}
\begin{figure}[ht]
    \centering
    \includegraphics[width=\linewidth]{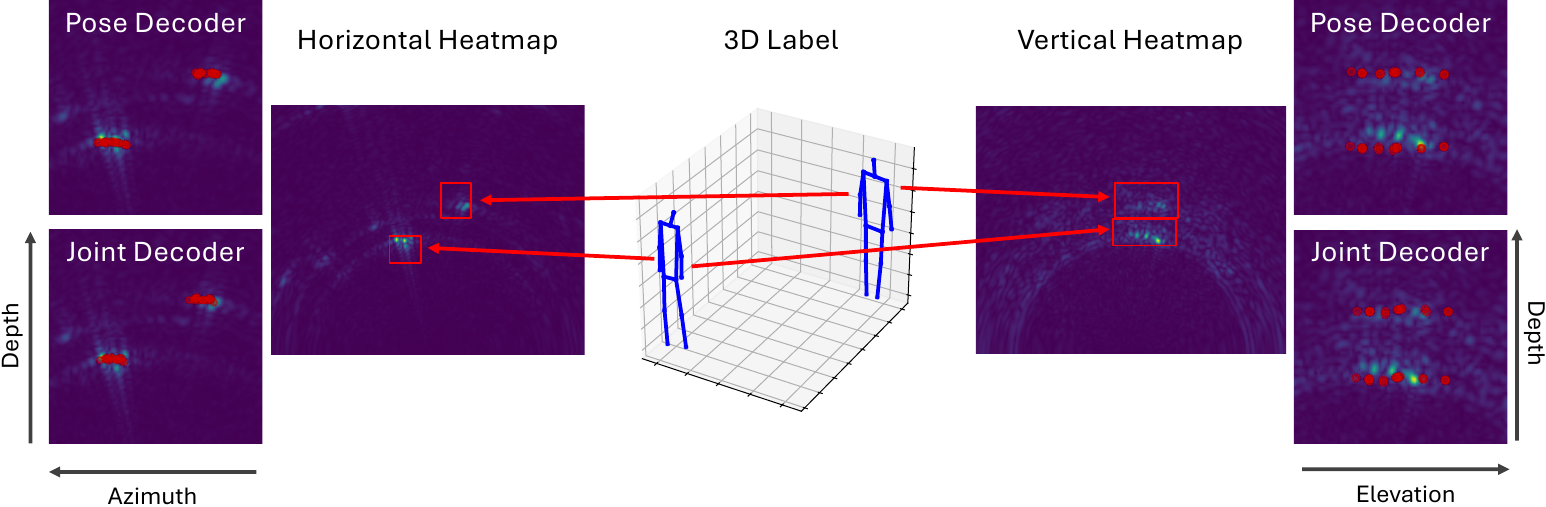}
    \caption{Visualization of pseudo-3D deformable attention. The attention mechanism samples features around the bright signals caused by body reflection, represented as red dots on the plots.}
    \label{fig:attention_vis}
\end{figure}

\paragraph{Visualization of Pseudo-3D Deformable Attention:}
Fig.~\ref{fig:attention_vis} shows the visualization of pseudo-3D deformable attention at the last layers in the pose and joint decoders.
For each view (horizontal on the left, vertical on the right), close-up regions around the bright radar reflections corresponding to subjects are extracted and visualized.
The red dots on the plots indicate the sampling locations selected by deformable attention.
The pseudo-3D deformable attention mechanism primarily samples features from regions surrounding human subjects.
Specifically, in the vertical view, the sampling points are clearly divided and distributed by joint.
Moreover, as the sampling offsets are computed across the $x$, $y$, and $z$ axes at once in our pseudo-3D deformable attention, the sampling locations maintain consistent alignment in the depth direction across the views for the same subject.

\paragraph{Additional Visualization Cases:}
\begin{figure}[p]
    \centering
    \begin{minipage}[b]{\hsize}
        \centering
        \includegraphics[width=\linewidth]{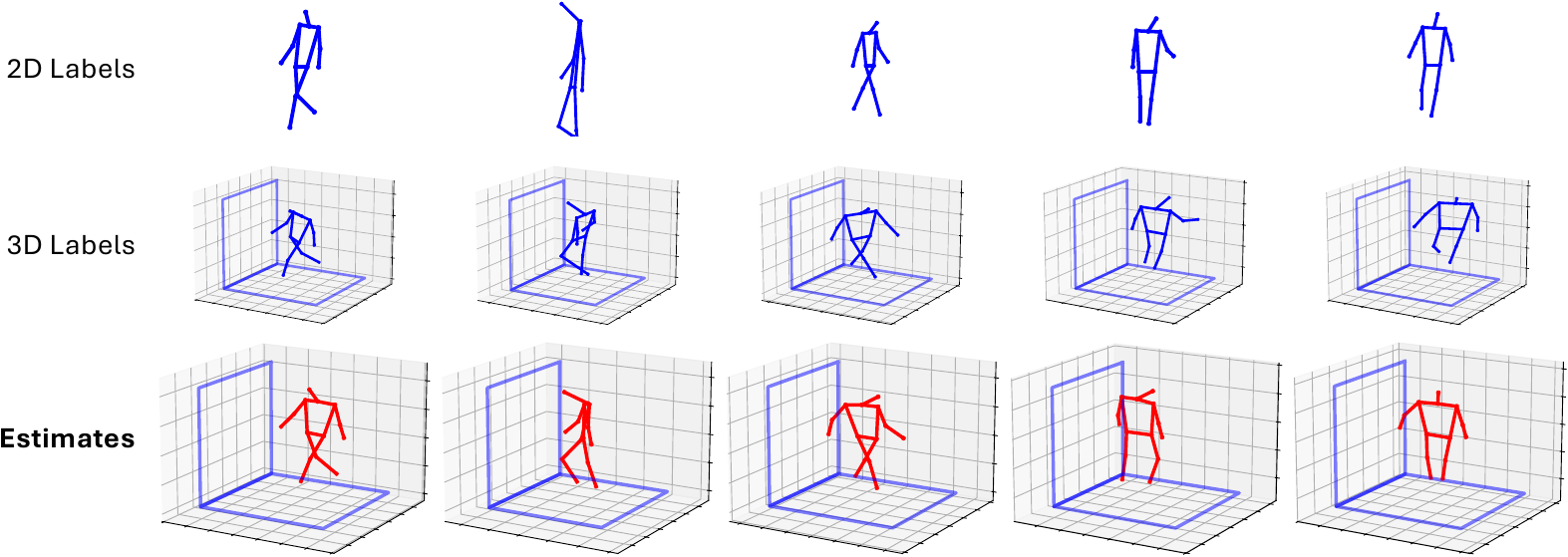}
        \subcaption{HIBER (WALK)}
    \end{minipage}
    \vspace{4mm}
    \begin{minipage}[b]{\hsize}
        \centering
        \includegraphics[width=\linewidth]{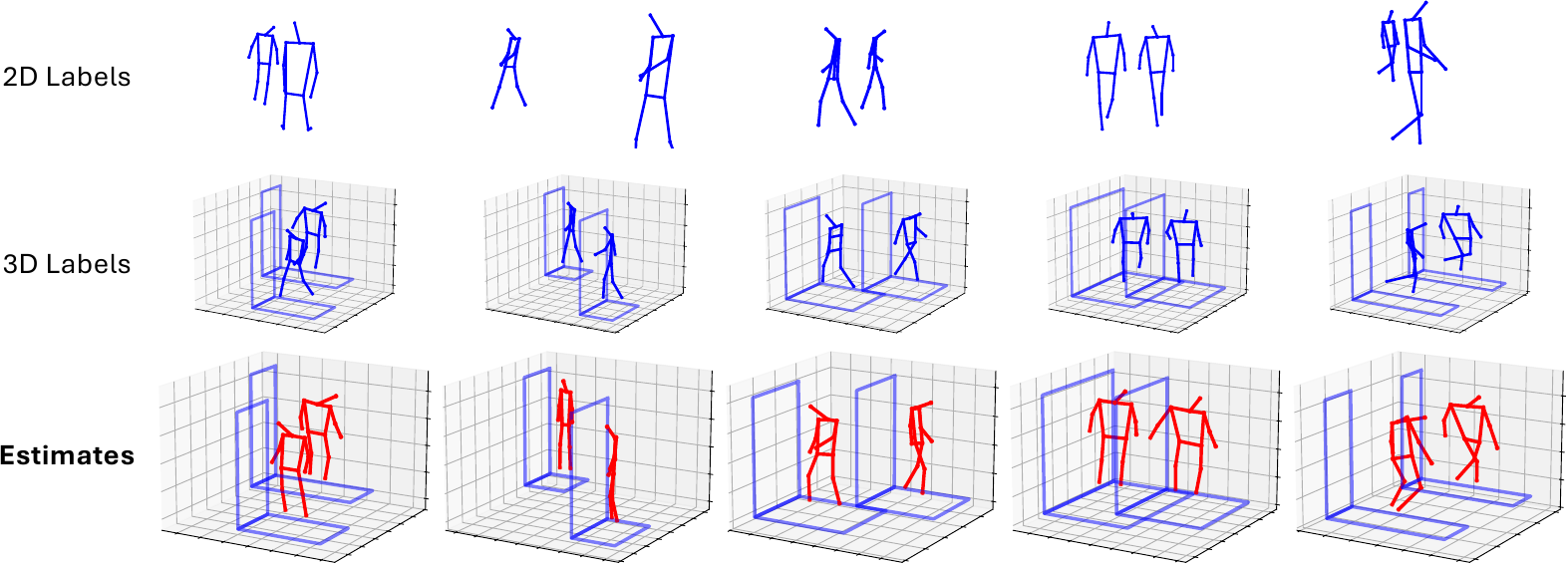}
        \subcaption{HIBER (MULTI)}
    \end{minipage}
    \vspace{4mm}
    \begin{minipage}[b]{\hsize}
        \centering
        \includegraphics[width=\linewidth]{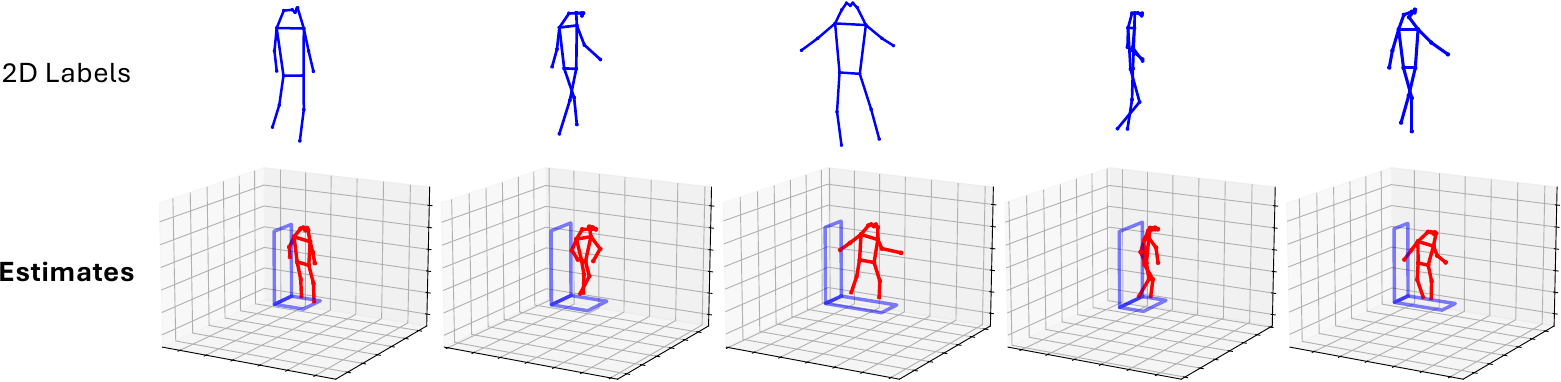}
        \subcaption{MMVR (P1S1)}
    \end{minipage}
    \vspace{4mm}
    \begin{minipage}[b]{\hsize}
        \centering
        \includegraphics[width=\linewidth]{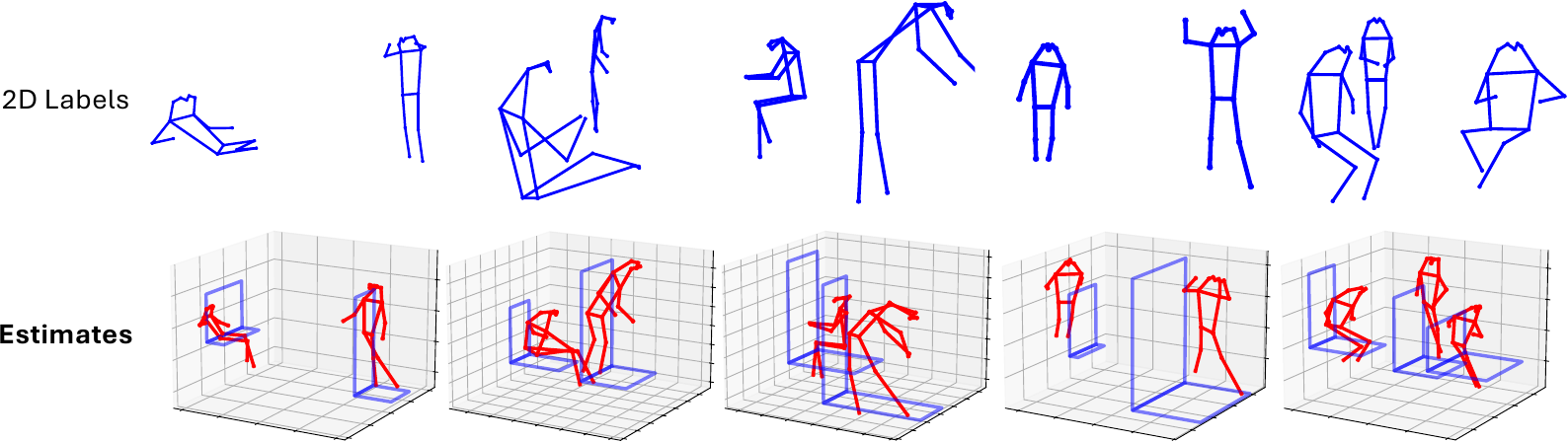}
        \subcaption{MMVR (P2S1)}
    \end{minipage}
    \caption{Visualization for RAPTR 3D pose estimation.}
    \label{fig:additional_visualization}
\end{figure}
We provide visualizations of the RAPTR estimation results for more cases in Fig.~\ref{fig:additional_visualization}.
We present visualizations of 2D and 3D labels and 3D pose estimates for HIBER WALK, MULTI, MMVR P1S1, and P2S1, arranged from top to bottom.

For the HIBER dataset, RAPTR maintains a stable estimation quality by capturing overall body orientation and limb articulation in (a), where only a single subject is presented, and (b) with multiple subjects.
Although minor inaccuracies are observed in some cases, RAPTR preserves the spatial arrangement and relative depth of the subjects.
As can be observed in the figure, the annotated 3D BBoxes are often significantly larger than the actual human body size, which originates from the dataset itself.
Since RAPTR utilizes only the center of the 3D BBoxes as reference for coarse-grained 3D cue, it remains largely unaffected by such inaccuracies in box scale.

For the MMVR dataset, RAPTR has to deal with a diverse range of body configurations, including seated and crouched poses, and increased subject variability.
In the P1S1 setting (c), RAPTR consistently estimates plausible 3D pose estimations that are well aligned with 2D keypoint labels, even for non-standard upright posture like spreading arms.
On the other hand, in the P2S1 setting (d), RAPTR demonstrates reasonable performance for more complex body configurations with multiple subjects: it often captures the overall structure and spatial arrangement of each individual.
In some cases, RAPTR even reconstructs plausible limb configurations where the 2D labels are inaccurate or incomplete, such as the seated person's legs in the leftmost example of (d).
This suggests that the architectural design, which first estimates initial poses using the pose decoder with a template pose, then refines joint positions via the joint decoder, effectively preserves a human-body prior throughout the process.
However, there are visible imperfections in some cases, such as over-extended limbs or inaccurate limb orientations, illustrating the difficulty of 3D pose estimation in scenarios with occlusion and extreme articulation.

\paragraph{Failure Cases:}
\begin{figure}[htb]
    \centering
    \includegraphics[width=\linewidth]{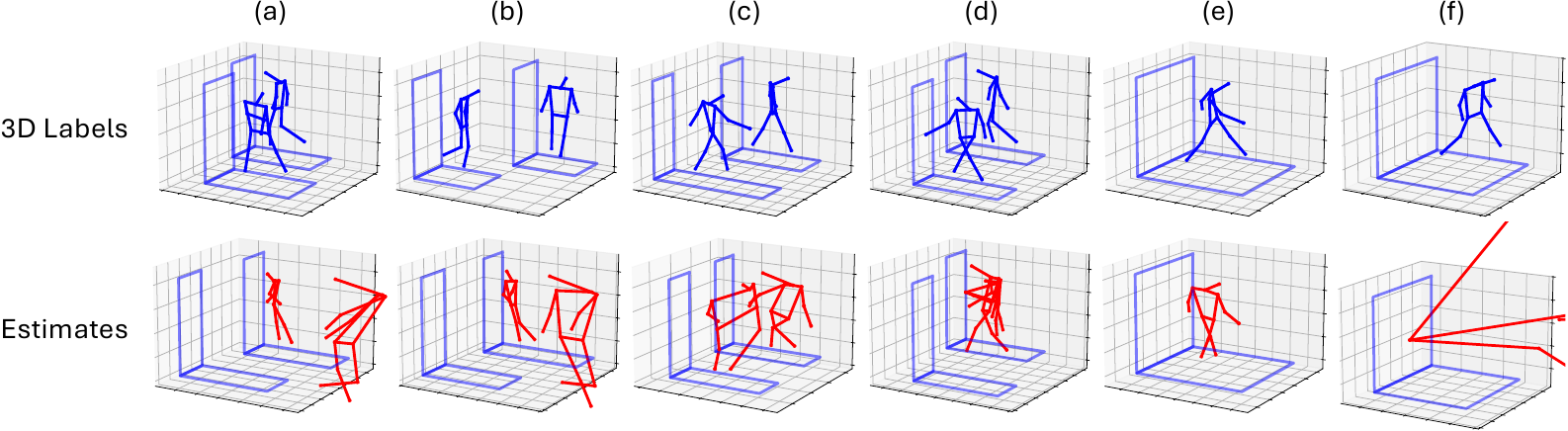}
    \caption{Failure cases for the HIBER (MULTI) dataset.}
    \label{fig:app_failure_cases}
\end{figure}
We provide failure cases for the HIBER MULTI dataset in Fig.~\ref{fig:app_failure_cases}.
In (a) and (b), RAPTR locates pose estimations significantly away from the 3D BBox labels, indicating errors in both subject localization and depth reasoning.
In (c), the estimated poses appear severely distorted, failing to preserve the anatomical structure of the body.
Despite being spatially close to the correct locations, the joint configurations are implausible, indicating a breakdown in fine-grained pose refinement.
In (d), RAPTR fails to correctly associate the estimated poses with the 3D BBoxes, leading to redundant estimates that multiple predictions are assigned to the same individual, degrading the quality of the final estimation.
In (e), RAPTR fails to recover the correct body orientation.
While the label pose is stepping forward with the left leg in the back-right direction, the estimated pose incorrectly keeps the body facing forward and places the right leg in an unnatural cross-step position.
In (f), RAPTR produces a completely corrupted estimate with no apparent correspondence to the human pose.
These failures are frequently observed when the two individuals overlap on the 2D image plane and the 2D keypoint labels themselves exhibit pose inaccuracies, indicating the limitation of our RAPTR, which relies on the 2D keypoint labels to accurately represent human shapes.

\end{document}